\documentclass[journal]{IEEEtran}
\usepackage{amsmath,amsfonts}
\usepackage{array}
\usepackage[caption=false,font=normalsize,labelfont=sf,textfont=sf]{subfig}
\usepackage{textcomp}
\usepackage{stfloats}
\usepackage{url}
\usepackage{verbatim}
\usepackage{graphicx}
\usepackage{cite}
\usepackage{bm}
\usepackage{amssymb}

\usepackage{booktabs}
\usepackage{hyperref}

\newtheorem{remark}{Remark}

\newtheorem{lemma}{Lemma}

\newtheorem{problem}{Problem}

\newtheorem{definition}{Definition}
\usepackage{xcolor}

\usepackage{enumitem}

\usepackage{tabularx}

\usepackage[caption=false]{subfig}

\usepackage{multirow}
\usepackage{threeparttable}

\usepackage{algorithm}
\usepackage{algpseudocode}

\definecolor{colorMSACL}{HTML}{0095FF}
\definecolor{colorSAC}{HTML}{FF7F0E}
\definecolor{colorLAC}{HTML}{2CA02C}
\definecolor{colorPPO}{HTML}{D62728}
\definecolor{colorPOLYC}{HTML}{4D3464}

\newcommand{\legenditem}[2]{{\color{#1}\rule[0pt]{10pt}{6pt}}\hspace{4pt}\textsf{#2}}

\usepackage[table]{xcolor}
\definecolor{tablegray}{RGB}{240,240,240}

\begin{document}
\title{MSACL: Multi-Step Actor-Critic Learning with Lyapunov Certificates for Exponentially Stabilizing Control}

\author{Yongwei Zhang, Yuanzhe Xing, Quanyi Liang, Quan Quan,~\IEEEmembership{Senior Member,~IEEE}, and Zhikun She
\thanks{Yongwei Zhang, Yuanzhe Xing and Quanyi Liang are with the School of Mathematical Sciences, Beihang University, Beijing 100191, China (e-mail: \{zhangyongwei, mathxyz, qyliang\}@buaa.edu.cn). \textit{These three authors contributed equally to this work.}}
\thanks{Quan Quan is with the School of Automation Science and Electrical Engineering, Beihang University, Beijing 100191, China (e-mail: qq\_buaa@buaa.edu.cn).}
\thanks{Zhikun She is with the School of Mathematical Sciences, Beihang University, Beijing 100191, China, and also with the Fujian Key Laboratory of Financial Information Processing, Putian University, Putian
351100, China (e-mail: zhikun.she@buaa.edu.cn).}
\thanks{Corresponding authors: Quan Quan and Zhikun She.}
\thanks{This work was supported by the National Key Research and Development
Program of China (No. 2022YFA1005103) and National Natural Science Foundation of China (NSFC 12371452).}
}



\maketitle

\begin{abstract}
For stabilizing control tasks, model-free reinforcement learning (RL) approaches face numerous challenges, particularly regarding the issues of effectiveness and efficiency in complex high-dimensional environments with limited training data. To address these challenges, we propose Multi-Step Actor-Critic Learning with Lyapunov Certificates (MSACL), a novel approach that integrates exponential stability into off-policy maximum entropy reinforcement learning (MERL). 
In contrast to existing RL-based approaches that depend on elaborate reward engineering and single-step constraints, MSACL adopts intuitive reward design and exploits multi-step samples to enable exploratory actor-critic learning. 
Specifically, we first introduce Exponential Stability Labels (ESLs) to categorize training samples and propose a $\lambda$-weighted aggregation mechanism to learn Lyapunov certificates. Based on these certificates, we further design a stability-aware advantage function to guide policy optimization, thereby promoting rapid Lyapunov descent and robust state convergence.
We evaluate MSACL across six benchmarks, comprising four stabilizing and two high-dimensional tracking tasks. Experimental results demonstrate its consistent performance improvements over both standard RL baselines and state-of-the-art Lyapunov-based RL algorithms. Beyond rapid convergence, MSACL exhibits robustness against environmental uncertainties and generalization to unseen reference signals. 
The source code and benchmarking environments are available at \href{https://github.com/YuanZhe-Xing/MSACL}{https://github.com/YuanZhe-Xing/MSACL}.

\end{abstract}


\section{Introduction}
\IEEEPARstart{S}{tabilizing} control is a central requirement in real-world control systems, especially in safety-critical applications such as autonomous driving \cite{zhang2024adaptive}, aerospace systems \cite{holt2025reinforcement}, and robotics \cite{khader2021learning}. In these settings, a controller must not only accomplish the desired task but also guarantee stable system behavior under uncertainty and disturbances \cite{choi2020reinforcement, wang2024safe}. As control systems become increasingly complex and are deployed in dynamic, partially unknown environments, developing control strategies that can simultaneously ensure stability and scalability has become a fundamental challenge \cite{brunke2022safe}.
Traditionally, control strategies have relied heavily on rigorous mathematical frameworks and precise analytical models\cite{lu2019stabilizability, wang2022inner}, such as constrained Markov Decision Processes (cMDPs) \cite{altman1999constrained}, Hamilton-Jacobi reachability analysis \cite{bansal2017hamilton} and certificate-based methods utilizing strict control-theoretic tools (e.g., sum-of-squares optimization \cite{giesl2015review, liang2025efficient} and algebraic synthesis \cite{she2014discovering}). However, all these traditional approaches face severe scalability challenges when applied to complex nonlinear systems with unknown dynamics.

To address the complexities of real-world dynamical systems without requiring explicit analytical models, RL has emerged as a highly effective paradigm for optimal control \cite{kober2013rl}, demonstrating considerable empirical success across various control tasks \cite{luckel2020probabilistic, zhu2017target}. Nevertheless, standard model-free RL algorithms, such as DDPG \cite{lillicrap2016continuous}, PPO \cite{schulman2017proximal}, and SAC \cite{haarnoja2018soft}, lack intrinsic stability awareness. They treat stability as an incidental byproduct of reward maximization, necessitating extensive reward engineering to achieve stable or safe behaviors. To bridge this gap, Lyapunov-based RL algorithms have been developed to integrate stability constraints into the learning process. For instance, Policy Optimization with Lyapunov Certificates (POLYC) \cite{chang2021stabilizing} incorporates Lyapunov certificates within the PPO framework, while Lyapunov-based Actor-Critic (LAC) \cite{han2020actor} introduces an approach for learning stability-constrained critics. By augmenting standard RL architectures with learned neural Lyapunov certificates, these methods empirically facilitate structural stability for nonlinear control systems.

However, despite these advances, existing Lyapunov-based RL methods often struggle with both effectiveness and efficiency in complex, high-dimensional stabilizing control tasks (e.g., Quadrotor tracking). The primary bottleneck for effectiveness lies in their inability to learn sufficiently accurate Lyapunov certificates. First, existing methods rely on single-step constraints, providing limited long-horizon information and leading to myopic policy improvement. Second, on-policy methods like POLYC depend exclusively on newly collected data, restricting the diversity of behaviors needed to accurately shape the Lyapunov landscape. Consequently, the learned certificates fail to provide reliable structural guidance, causing performance degradation or divergence.

To overcome the effectiveness issue, introducing a multi-step off-policy approach is essential, as it captures long-horizon trajectory evolution. Meanwhile, the multi-step off-policy approach naturally resolves the secondary issue: it boosts efficiency by deeply reusing historical experiences, overcoming the low sample efficiency inherent in previous on-policy methods.
Inspired by the idea of using multi-step off-policy, we in this paper attempt to learn a high-fidelity Lyapunov certificate that can reliably guide policy updates.
To this end, we propose Multi-Step Actor-Critic Learning with Lyapunov Certificates (MSACL). First, MSACL exploits multi-step samples to learn Lyapunov certificates, enabling the certificates to capture long-horizon system dynamics rather than relying only on local single-step transitions. Second, MSACL is built upon an off-policy maximum entropy reinforcement learning (MERL) framework, which not only improves sample efficiency through data reuse but also promotes broader exploration, thereby providing more diverse training data for characterizing system behaviors. While parameterizing Lyapunov certificates via neural networks inherently lacks the strict formal mathematical guarantees of traditional analytical methods, the proposed multi-step and off-policy learning paradigm yields a more reliable approximation empirically. This enhanced Lyapunov certificate, in turn, provides structurally grounded, stability-aware advantage estimates to guide policy optimization toward rapid and robust stabilization.

The proposed MSACL approach makes the following contributions to safe model-free control:

\begin{enumerate}[label=(\roman*), noitemsep]
\item \textbf{Multi-Step Learning with Lyapunov Guidance.} We introduce a multi-step learning mechanism to learn high-precision Lyapunov certificates that better capture long-horizon system dynamics and provide more effective guidance for policy updates. Built upon this mechanism, MSACL further incorporates exponential stability through Exponential Stability Labels (ESLs) and a $\lambda$-weighted aggregation scheme, yielding stronger stability supervision and faster convergence than conventional single-step methods.

\item \textbf{Sample-Efficient Off-Policy Framework.} We develop MSACL within an off-policy MERL framework, moving beyond the data inefficiency of existing on-policy Lyapunov-based methods. By improving data reuse and encouraging broader exploration, the proposed framework captures more diverse system behaviors and provides richer training samples for Lyapunov certificate learning. In addition, importance sampling is incorporated to mitigate distribution shifts, further enhancing sample efficiency and training stability.

\item \textbf{Systematic Evaluation and Robustness Validation.} We establish a comprehensive evaluation platform across six benchmarks, including high-dimensional stabilizing control tasks. Experimental results demonstrate MSACL's consistent empirical superiority over both standard RL and state-of-the-art Lyapunov-based baselines. Furthermore, MSACL exhibits robustness against environmental noise and generalization to unseen reference signals.

\end{enumerate}

The remainder of this article is organized as follows. Section \ref{sec:Preliminaries} formulates the control problems and establishes necessary theoretical foundations. Section \ref{sec:lyapunov} describes the Lyapunov certificate learning methodology. Section \ref{sec:policy} presents the optimization process within the actor-critic framework. Section \ref{sec:experiments} provides comprehensive experimental evaluations across six benchmarks. Finally, Section \ref{sec:conclusion} discusses the design philosophy and concludes this article.

\section{Preliminaries} \label{sec:Preliminaries}
This section establishes the theoretical foundations for MSACL. We first define the exponentially stabilizing control problems for model-free discrete-time systems, then introduce Lyapunov-based exponential stability criteria, and finally review the MERL framework.

\subsection{Problem Statement}
In this article, we study the stabilizing problem for general discrete-time deterministic control systems:
\begin{equation}\label{control_system}
    \mathbf{x}_{t+1} = f(\mathbf{x}_t, \mathbf{u}_t),
\end{equation}
where $\mathbf{x}_t \in \mathcal{X} \subseteq \mathbb{R}^d$ denotes the state at time $t$, $\mathbf{u}_t \in \mathcal{U} \subseteq \mathbb{R}^m$ denotes the control input, $\mathcal{X}$ and $\mathcal{U}$ represent the sets of admissible states and control inputs respectively, and $f: \mathcal{X} \times \mathcal{U} \to \mathcal{X}$ is the flow map. Given that $f$ is unknown, we adopt a model-free control paradigm. 
Let $\mathcal{X}_0 \subseteq \mathcal{X}$ denote the set of initial states. For any $\mathbf{x}_0 \in \mathcal{X}_0$, let $\mathbf{x}(t, \mathbf{x}_0)$ denote the state at time $t$ along the trajectory starting from $\mathbf{x}_0$. For notational simplicity, $\mathbf{x}_t$ is used to denote $\mathbf{x}(t, \mathbf{x}_0)$ whenever the context is unambiguous.
It is worth noting that the discrete-time case presented in this article can be similarly extended to the continuous-time case; interested readers may refer to \cite{dawson2023safe} and \cite{chang2021stabilizing} for details.

For the control system (\ref{control_system}), our objective is to find a state feedback policy $\pi: \mathcal{X} \to \mathcal{U}$ such that the control input $\mathbf{u}_t = \pi(\mathbf{x}_t)$ renders the closed-loop system $\mathbf{x}_{t+1} = f(\mathbf{x}_t, \pi(\mathbf{x}_t))$ to satisfy exponential stability. For clarity, we consider the scenario where the state $\mathbf{x}_t$ is fully observable, and the policy $\pi$ is stationary, i.e., it depends only on the state and not explicitly on time. Specifically, the problem studied in this article is formulated as:

\begin{problem}\label{problem1}
    \textit{(Exponential stabilizing control)} For the control system (\ref{control_system}), design a policy $\pi$ such that for any initial state $\mathbf{x}_0 \in \mathcal{X}_0$, the resulting closed-loop trajectory $\mathbf{x}(t, \mathbf{x}_0)$ exponentially converges to the equilibrium state $\mathbf{x}_g$.
\end{problem}

Without loss of generality, let $\mathbf{x}_g$ denote the origin, i.e., $\mathbf{x}_g = \mathbf{0}$. Furthermore, the tracking problem is inherently encompassed within the stabilizing framework, as tracking a time-varying reference signal $\mathbf{x}_t^{\text{ref}}$ reduces to stabilizing the error $\mathbf{e}_t = \mathbf{x}_t - \mathbf{x}_t^{\text{ref}}$ at zero.

\subsection{Exponential Stability and Lyapunov Certificates}
\begin{definition}\label{def1}
For the control system (\ref{control_system}), the equilibrium state $\mathbf{x}_g$ is said to be exponentially stable if there exist constants $C > 0$ and $0 < \eta < 1$ such that
\begin{equation}
\|\mathbf{x}_t - \mathbf{x}_g\| \leq C \eta^t \|\mathbf{x}_0 - \mathbf{x}_g\|, \quad \forall \mathbf{x}_0 \in \mathcal{X}_0, \forall t \ge 0.
\end{equation}
\end{definition}

To facilitate that the closed-loop system is exponentially stable with respect to $\mathbf{x}_g$, we introduce the following lemma:
\begin{lemma}\label{lemma1}
For the control system (\ref{control_system}), if there exists a continuous function $V: \mathcal{X} \mapsto \mathbb{R}$ and positive constants $\alpha_1, \alpha_2 > 0$ and $0 < \alpha_3 < 1$ satisfying:
\begin{equation}\label{thm_con}
    \begin{aligned}
    \alpha_1 \|\mathbf{x}_t\|^2 \leq V(\mathbf{x}_t) &\leq \alpha_2 \|\mathbf{x}_t\|^2, \\
    V(\mathbf{x}_{t+1}) - V(\mathbf{x}_t) &\leq -\alpha_3 V(\mathbf{x}_t),
    \end{aligned}
\end{equation}
for all $t \in \mathbb{Z}_{\geq 0}$, where $\mathbf{x}_{t+1} = f(\mathbf{x}_t, \mathbf{u}_t)$, then the origin $\mathbf{x}_g = \mathbf{0}$ is exponentially stable. The function $V$ is referred to as an exponentially stabilizing control Lyapunov function.
\end{lemma}

Lemma \ref{lemma1} is a classic result in control theory, while its continuous-time counterpart is discussed in \cite{ames2014rapidly}, we provide a concise derivation of the discrete-time version to elucidate its underlying mechanism. 

Specifically, starting from the descent condition $V(\mathbf{x}_{t+1}) \leq (1 - \alpha_3) V(\mathbf{x}_t)$ in (\ref{thm_con}), the multi-step relationship can be established iteratively as:
\begin{equation}\label{multi-step}
    V(\mathbf{x}_{t+n}) \leq (1 - \alpha_3) V(\mathbf{x}_{t+n-1}) \leq \dots \leq (1 - \alpha_3)^n V(\mathbf{x}_t).
\end{equation}
By incorporating the lower and upper quadratic bounds, i.e., $\alpha_1 \|\mathbf{x}_{t+n}\|^2 \leq V(\mathbf{x}_{t+n})$ and $ V(\mathbf{x}_t) \leq \alpha_2 \|\mathbf{x}_t\|^2$, the relationship between state norms across multiple steps follows as:
\begin{equation}\label{multi-step_conclusion}
    \|\mathbf{x}_{t+n}\| \leq \sqrt{\frac{\alpha_2}{\alpha_1}} (1 - \alpha_3)^{n/2} \|\mathbf{x}_t\|.
\end{equation}
Furthermore, by setting the initial time to $t = 0$ and denoting the current time step as $n = t \in \mathbb{Z}_{\geq 0}$, we obtain the global trajectory bound:
\begin{equation*}
    \|\mathbf{x}_t\| \leq \sqrt{\frac{\alpha_2}{\alpha_1}} (1 - \alpha_3)^{t/2} \|\mathbf{x}_0\|.
\end{equation*}
By defining the constants $C = \sqrt{\alpha_2 / \alpha_1}$ and $\eta = \sqrt{1 - \alpha_3}$, the result satisfies Definition \ref{def1}, thereby proving that the control system (\ref{control_system}) is exponentially stable with respect to $\mathbf{x}_g = \mathbf{0}$.

\subsection{Standard RL and Maximum Entropy RL}
To begin the analysis, we interpret the control system (\ref{control_system}) as a discrete-time environment with continuous state space $\mathcal{X}$ and action space $\mathcal{U}$ under the standard RL paradigm. At each time step $t$, the RL agent observes the state $\mathbf{x}_t \in \mathcal{X}$ and executes an action $\mathbf{u}_t \in \mathcal{U}$, causing the environment to transition to the next state $\mathbf{x}_{t+1}$ and yield a scalar reward $r_t = r(\mathbf{x}_t, \mathbf{u}_t)$. This reward is formulated to penalize state deviations from the target $\mathbf{x}_g$ (or reference signal $\mathbf{x}_t^{\text{ref}}$) while accounting for the associated control effort. 

Furthermore, we employ stochastic policies in this article to facilitate exploration within the MERL framework. A stochastic policy is defined as a mapping $\pi: \mathcal{X} \to \mathcal{P}(\mathcal{U})$, where $\mathcal{P}(\mathcal{U})$ denotes the set of probability distributions over $\mathcal{U}$, and $\mathbf{u}_t \sim \pi(\cdot|\mathbf{x}_t)$ is sampled from the probability distribution. We denote the state and state-action marginals induced by $\pi$ as $\rho_\pi(\mathbf{x}_t)$ and $\rho_\pi(\mathbf{x}_t, \mathbf{u}_t)$, respectively. In contrast to standard RL, which maximizes the expected cumulative discounted reward, MERL optimizes an entropy-augmented objective \cite{haarnoja2018soft, duan2022distributional, duan2025distributional}:
\begin{equation}\label{J_pi}
    J_{\pi} = \underset{(\mathbf{x}_{i \geq t}, \mathbf{u}_{i \geq t}) \sim \rho_{\pi}}{\mathbb{E}} \left[ \sum_{i=t}^{\infty} \gamma^{i-t} \left[ r_{i} + \alpha \mathcal{H}(\pi(\cdot|\mathbf{x}_{i})) \right] \right],
\end{equation}
where $\gamma \in [0, 1)$ is the discount factor, $\alpha > 0$ is the temperature coefficient, and $\mathcal{H}(\pi(\cdot|\mathbf{x})) = \mathbb{E}_{\mathbf{u} \sim \pi(\cdot|\mathbf{x})} [-\log \pi(\mathbf{u}|\mathbf{x})]$ is the policy entropy.

The corresponding soft Q-function, $Q^\pi(\mathbf{x}_t, \mathbf{u}_t)$, represents the expected soft return, which comprises both rewards and entropy terms. The soft Bellman operator $\mathcal{T}^\pi$ is defined as:
\begin{equation}\label{Bellman_operator}
    \begin{aligned}
        \mathcal{T}^{\pi}Q^{\pi}(\mathbf{x}_{t}, \mathbf{u}_{t}) \triangleq r_t + \gamma \underset{\mathbf{u}_{t+1} \sim \pi}{\mathbb{E}} \Big[ &Q^{\pi}(\mathbf{x}_{t+1}, \mathbf{u}_{t+1}) \\ 
        & - \alpha \log \pi(\mathbf{u}_{t+1}|\mathbf{x}_{t+1}) \Big],
    \end{aligned}
\end{equation}
where $\mathbf{x}_{t+1} = f(\mathbf{x}_t, \mathbf{u}_t)$. In the soft policy iteration (SPI) algorithm \cite{haarnoja2018soft}, the policy is optimized by alternating between soft policy evaluation, where $Q^\pi$ is updated via $\mathcal{T}^\pi$, and soft policy improvement, which is equivalent to finding  a policy $\pi$ that maximizes the soft Q-function for $\mathbf{x}_t$:
\begin{equation}\label{max_softQ}
    \pi_{\rm new} = \arg\max_{\pi} \underset{\mathbf{x}_t \sim \rho_\pi, \mathbf{u}_t \sim \pi}{\mathbb{E}} \left[ Q^{\pi_{\rm old}}(\mathbf{x}_t, \mathbf{u}_t) - \alpha \log \pi(\mathbf{u}_t|\mathbf{x}_t) \right].
\end{equation}

Although SPI converges to the optimal policy $\pi^*$ in tabular settings, the control problems studied in this article involve continuous state and action spaces necessitating function approximation. Thus, we parameterize the soft Q-function $Q_\theta(\mathbf{x}_t, \mathbf{u}_t)$ and the policy $\pi_\phi(\mathbf{u}_t|\mathbf{x}_t)$ using neural networks with parameters $\theta$ and $\phi$. We assume $\pi_\phi$ follows a diagonal Gaussian distribution, and actions are sampled using the reparameterization trick \cite{haarnoja2018soft}: $\mathbf{u}_t = g_\phi(\mathbf{x}_t; \epsilon_t)$, where $\epsilon_t \sim \mathcal{N}(\mathbf{0}, \mathbf{I}_m)$, ensuring that the optimization objectives remain differentiable with respect to $\phi$.

\section{Learning Lyapunov Certificate}\label{sec:lyapunov}
This section details the model-free Lyapunov certificate learning mechanism. We first introduce a sliding-window strategy for multi-step data collection, and then design a loss function for the Lyapunov network (also referred to as the Lyapunov certificate), incorporating both boundedness and stability constraints.

\subsection{Data Collection}
In contrast to conventional Lyapunov-based RL algorithms \cite{chang2021stabilizing, han2020actor}, which primarily target asymptotic stability through single-step transitions, our proposed MSACL enforces the stricter exponential stability requirements. Guided by the multi-step relationship (\ref{multi-step}), establishing exponential stability necessitates leveraging multi-step sequence samples from state trajectories. This sequential data structure provides a sufficient temporal horizon to effectively learn the contraction properties of Lyapunov certificates.

To operationalize this, we define the fundamental transition tuple as $\mathbf{d}_t = (\mathbf{x}_t, \mathbf{u}_t, r_t, \pi_\phi(\mathbf{u}_t|\mathbf{x}_t), \mathbf{x}_{t+1})$. Unlike standard replay buffers that store these tuples individually, we employ a double-ended queue (deque) as a sliding window to capture temporal correlations. Upon reaching a length of $n$, the complete multi-step sequence $\{\mathbf{d}_t, \mathbf{d}_{t+1}, \dots, \mathbf{d}_{t+n-1}\}$ is packaged and transferred to the replay buffer $\mathcal{D}$, while the oldest entry is concurrently removed to maintain the window size.

\begin{figure}[!t] 
	\centering
	\includegraphics[width = 3.0in]{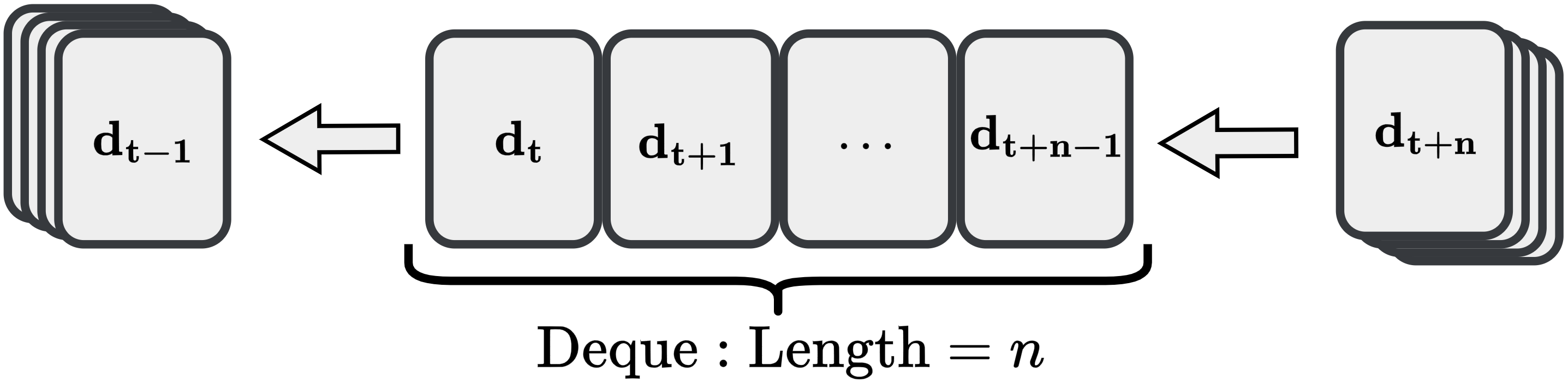}
	\caption{Schematic of $n$-step sliding window data collection. A double-ended queue (deque) of size $n$ buffers consecutive transitions $\mathbf{d}_t$ to provide complete multi-step trajectories.}
	\label{data_sample}
\end{figure}

\subsection{Loss Function Design of Lyapunov Network}\label{sec:lyapunov_loss}
Similar to the soft Q-network and policy network, we introduce a Lyapunov network $V_\psi$ parameterized by $\psi$ to estimate the Lyapunov certificate. In designing the loss function, We strictly adhere to the boundedness and multi-step stability conditions in Lemma \ref{lemma1}.

\textbf{(1) Boundedness Loss}.
Based on the boundedness condition in (\ref{thm_con}), we construct the boundedness loss function to ensure the Lyapunov certificate to be quadratically constrained:
\begin{equation}
\begin{aligned}
    \mathcal{L}_{\text{Bnd}}(\psi) = \frac{1}{Nn} & \sum_{i=1}^{N}\sum_{k=0}^{n-1} \Big[ \max\left( \alpha_1 \|\mathbf{x}_{t+k}^i\|^2 - V_\psi(\mathbf{x}_{t+k}^i), 0 \right) \\
    & + \max\left( V_\psi(\mathbf{x}_{t+k}^i) - \alpha_2 \|\mathbf{x}_{t+k}^i\|^2, 0 \right)\Big],
\end{aligned}
\end{equation}
where $N$ denotes the batch size of multi-step sequence samples, $i$ denotes the $i$-th sample, and $n$ is the sequence length. This loss enforces the boundedness condition on $V_\psi$.

\textbf{(2) Multi-step Stability Loss}.
A major challenge in learning Lyapunov certificates from off-policy data is the discrepancy between the behavior policy $\pi_{\phi_{\text{old}}}$ and the target policy $\pi_{\phi}$. To mitigate this discrepancy, we employ a sequence-wise importance sampling strategy.
Specifically, for a multi-step sequence $\{\mathbf{d}_t,\dots,\mathbf{d}_{t+n-1}\}$ generated by $\pi_{\phi_{\text{old}}}$, we compute the cumulative importance sampling ratio for the transition from $t$ to $t+k$ ($k=1,\dots,n-1$) as:
\begin{equation}\label{IS}
    \text{IS}_{\text{clip},k} = \prod_{j=t}^{t+k-1} \min \left( \frac{\pi_{\phi}(\mathbf{u}_j|\mathbf{x}_j)}{\pi_{\phi_{\text{old}}}(\mathbf{u}_j|\mathbf{x}_j)}, 1 \right).
\end{equation}
Note that we apply a clipping operation to the importance sampling ratios to prevent variance explosion and enhance training stability. 

On the other hand, learning Lyapunov certificates directly from off-policy data may not be accurate, as empirical sequences in the early stages of training  often fail to satisfy the exponential decay condition (\ref{multi-step_conclusion}), leading to unreliable certificate estimation. To establish a strong empirical alignment between empirical data and theoretical stability requirements, we propose the Exponential Stability Label (ESL) mechanism. Specifically, we introduce \textit{positive} and \textit{negative} samples to bridge the gap between theory and practice:

\begin{definition}\label{p_n_sample}
    Within an $n$-step sequence sample, for the state $\mathbf{x}_{t+k}$, if it satisfies the exponential decay condition:
    \begin{equation*}
        \|\mathbf{x}_{t+k}\| \leq \sqrt{\frac{\alpha_2}{\alpha_1}} (1 - \alpha_3)^{k/2} \|\mathbf{x}_t\|,
    \end{equation*}
    then $\mathbf{d}_{t+k}$ is classified as a \textit{positive sample}; otherwise, it is classified as a \textit{negative sample}.
\end{definition}

Accordingly, we define the Exponential Stability Label as:
\begin{equation*}
    \text{ESL}_{k} = \begin{cases}
    1, & \text{if } \|\mathbf{x}_{t+k}\| \leq \sqrt{\frac{\alpha_2}{\alpha_1}} (1 - \alpha_3)^{k/2} \|\mathbf{x}_t\| \\
    -1, & \text{otherwise}.
    \end{cases}
\end{equation*}
This labeling mechanism allows the Lyapunov network to distinguish between transitions that contribute to system stability and those that violate the certificate requirements.

Furthermore, to balance the bias-variance trade-off in multi-step sequences, we introduce a weighting factor $\lambda \in (0, 1)$, inspired by the (truncated) $\lambda$-return in classic RL literature \cite{sutton1998reinforcement}. 
Specifically, for the stability condition $V_\psi(\mathbf{x}_{t+k}) \leq (1-\alpha_3)^{k} V_\psi(\mathbf{x}_t)$ from (\ref{multi-step}), small $k$ determines a short horizon with limited information. Given that the Lyapunov network is randomly initialized and potentially inaccurate in the early stages of training, relying on short-horizon data may introduce high estimation bias.
Conversely, large $k$ captures extended information to reduce bias but accumulates environmental stochasticity into higher variance. To this end, we employ the weighting factor $\lambda$ to aggregate multi-step information and balance bias-variance trade-off.

Finally, by integrating the positive/negative sample labels with the weighting factor $\lambda$, we construct the Lyapunov stability loss to enforce a rigorous correspondence between empirical samples and the stability condition (\ref{multi-step}):
\begin{equation}\label{loss_diff_V}
    \mathcal{L}_{\text{Stab}}(\psi) = \frac{1}{N} \sum_{i=1}^N \left[ \sum_{k=1}^{n-1} \frac{\lambda^{k-1} \mathcal{L}_{\text{diff}, k}^{i}(\psi)}{\sum_{j=1}^{n-1} \lambda^{j-1}} \right],
\end{equation}
where the step-wise difference loss $\mathcal{L}_{\text{diff}, k}^{i}(\psi)$ is defined as:
\begin{equation*}
\begin{aligned}
    \mathcal{L}_{\text{diff}, k}^{i}(\psi) = \,&\text{IS}_{\text{clip},k}^i \cdot \max\Big(0, \\
    &\text{ESL}_{k}^i \cdot \left(V_\psi(\mathbf{x}_{t+k}^i) - (1-\alpha_3)^{k} V_\psi(\mathbf{x}_t^i)\right)\Big).
\end{aligned}
\end{equation*}

\textbf{(3) Combined Lyapunov Loss}.
By integrating the boundedness loss and the multi-step stability loss, the comprehensive loss function for optimizing the Lyapunov network $V_\psi$ is formulated as:
\begin{equation}\label{loss_lya}
    \mathcal{L}_{\text{Lya}}(\psi) = \omega_{\text{Bnd}} \mathcal{L}_{\text{Bnd}}(\psi) + \omega_{\text{Stab}} \mathcal{L}_{\text{Stab}}(\psi),
\end{equation}
where $\omega_{\text{Bnd}}$ and $\omega_{\text{Stab}}$ are positive weighting coefficients.

\section{Learning Policy with Exponential Stability Constraints}\label{sec:policy}
This section details the MSACL optimization framework. We first establish soft Q-network updates using multi-step Bellman residuals to enhance data efficiency. Then we introduce a novel policy loss incorporating Lyapunov-based stability advantages with PPO-style clipping for robust exponential stabilization. Finally, we describe automated temperature adjustment and the complete actor-critic procedure.

\subsection{Loss Function Design of Soft Q Networks}
Following the soft Bellman operator (\ref{Bellman_operator}), we define the sampled Bellman residual for the $i$-th multi-step sequence sample of $Q_\theta$ as:
\begin{equation}
\begin{aligned}
    \mathcal{B}_t^{i}(\theta) = Q_\theta(\mathbf{x}_t^i, \mathbf{u}_t^i) - \Big[ & r_t^i + \gamma Q_{\bar{\theta}}(\mathbf{x}_{t+1}^i, \mathbf{u}_{t+1}^i) \\
    & - \alpha \log \pi_{\phi}(\mathbf{u}_{t+1}^i|\mathbf{x}_{t+1}^i) \Big],
\end{aligned}
\end{equation}
where $Q_{\bar{\theta}}$ denotes the target network parameterized by $\bar{\theta}$. Utilizing collected multi-step sequence samples, the loss function for the soft Q-network is defined as the mean squared Bellman error (MSBE) across the entire sequence:
\begin{equation}\label{loss_softQ}
    \mathcal{L}_{\text{SoftQ}}(\theta) = \frac{1}{Nn} \sum_{i=1}^{N} \sum_{k=0}^{n-1} (\mathcal{B}_{t+k}^{i}(\theta))^2.
\end{equation}

In contrast to standard MERL algorithms such as SAC \cite{haarnoja2018soft}, which typically utilize single-step transition (i.e., $n=1$), our approach leverages the complete $n$-step sequence to minimize the following objective function:
\begin{equation}\label{objective_function}
    \begin{aligned}
        \mathcal{J}_{\text{SoftQ}}(\theta)=&\underset{\mathbf{x}_t \sim \rho_\pi, \mathbf{u}_t \sim \pi}{\mathbb{E}}\Big[ Q_\theta(\mathbf{x}_t, \mathbf{u}_t) -  \Big( r_t + \gamma \underset{\mathbf{x}_{t+1} \sim \rho_\pi, \mathbf{u}_{t+1} \sim \pi}{\mathbb{E}}\\
        &\big(Q_{\bar{\theta}}(\mathbf{x}_{t+1}, \mathbf{u}_{t+1}) - \alpha \log \pi_{\phi}(\mathbf{u}_{t+1}|\mathbf{x}_{t+1})\big)  \Big) \Big]^2.
    \end{aligned}
\end{equation}

\begin{remark}
    Although minimizing (\ref{objective_function}) theoretically necessitates independent and identically distributed (i.i.d.) sampling for unbiased gradient estimation \cite{zhao2025mathematical}, our proposed multi-step approach strategically balances mathematical rigor with enhanced sample efficiency. Indeed, while $n$-step sequences introduce temporal correlations that deviate from strict i.i.d. assumptions, experimental evidence demonstrates that moderate sequence lengths maintain learning stability. Ultimately, the substantial gains in batch data information outweigh the minor bias introduced by short-term correlations, leading to accelerated convergence and more robust control performance.
\end{remark}

\subsection{Loss Function Design of Policy Network}
This subsection details the policy network optimization, which integrates the maximum entropy objective with a novel Lyapunov-based stability guidance mechanism.

 \textbf{(1) Soft-Q Value Maximization}.
Following the policy optimization (\ref{max_softQ}) in SPI, the first objective is to maximize the soft-Q value. For the $i$-th multi-step sequence sample $\{\mathbf{d}_t^i, \dots, \mathbf{d}_{t+n-1}^i\}$, the SAC-style component of the loss function is expressed as:
\begin{equation*}
    \begin{aligned}
        \mathcal{L}_{\pi,\text{SAC}}^{i}(\phi) = \frac{1}{n} \sum_{k=0}^{n-1} \Big[ & Q_\theta(\mathbf{x}_{t+k}^i, g_\phi(\mathbf{x}_{t+k}^i;\epsilon_{t+k}^i)) \\
        & - \alpha \log \pi_\phi(g_\phi(\mathbf{x}_{t+k}^i;\epsilon_{t+k}^i) | \mathbf{x}_{t+k}^i) \Big].
    \end{aligned}
\end{equation*}
Note that $g_\phi(\mathbf{x}_{t+k}^i;\epsilon_{t+k}^i)$ is obtained via the reparameterization trick \cite{haarnoja2018soft}.

\textbf{(2) Stability-Aware Advantage Guidance}.
To ensure exponential stability, we attempt to utilize the learned Lyapunov certificate $V_\psi$ to guide policy updates. Inspired by the Generalized Advantage Estimation (GAE) in PPO \cite{schulman2017proximal}, we transform the Lyapunov descent property into an active guidance signal. Specifically, we define the \textit{stability advantage function} based on $V_\psi$ at $t+k$ as a metric for convergence performance:
\begin{equation}
    A_{t,k} = (1-\alpha_3)^k V_\psi(\mathbf{x}_t) - V_\psi(\mathbf{x}_{t+k}).
\end{equation}

A positive stability advantage ($A_{t,k} \geq 0$) indicates that the system trajectory adheres to the exponential decay condition (\ref{multi-step}). Maximizing this advantage encourages actions that accelerate $V_\psi$ decay, thereby increasing the state convergence rate toward $\mathbf{x}_g$. Conversely, $A_{t,k} < 0$ signifies a certificate condition violation that the policy must learn to avoid. This stability-aware optimization aligns with recent developments in stable model-free control, such as D-learning \cite{quan2024control, shen2025dopt, liu2025dlclip}. Similar to (\ref{loss_diff_V}), in order to balance the bias-variance trade-off across varying horizons, we apply the weighting factor $\lambda$ to obtain the aggregated stability advantage:
\begin{equation}\label{aggregated_adv}
    A_{t,\lambda} = \sum_{k=1}^{n-1} \frac{\lambda^{k-1} A_{t,k}}{\sum_{j=1}^{n-1} \lambda^{j-1}}.
\end{equation}

 \textbf{(3) Integrated Policy Optimization Objective}.
For the $i$-th multi-step sequence sample, we define the first-step importance sampling (IS) ratio as $\rho^i(\phi) = \frac{\pi_\phi(\mathbf{u}_t^i|\mathbf{x}_t^i)}{\pi_{\phi_{\text{old}}}(\mathbf{u}_t^i|\mathbf{x}_t^i)}$. By incorporating the aggregated stability advantage (\ref{aggregated_adv}) with PPO-style clipping to prevent excessively large updates, the policy loss function is formulated as:
\begin{equation}\label{loss_policy}
\begin{aligned}
    \mathcal{L}_\pi(\phi) = -\frac{1}{N} \sum_{i=1}^N \Big[ & \mathcal{L}_{\pi,\text{SAC}}^{i}(\phi) + \min\Big(\rho^i(\phi) A_{t,\lambda}^i, \\
    & \text{clip}(\rho^i(\phi), 1-\varepsilon, 1+\varepsilon) A_{t,\lambda}^i \Big) \Big].
\end{aligned}
\end{equation}

\begin{remark}
    Notably, we employ distinct importance sampling strategies for the critic and the actor.
    In (\ref{loss_diff_V}), sequence-level IS provides low-bias correction for learning $V_\psi$ from off-policy data. This variance remains manageable via truncation in (\ref{IS}) since policy parameters are fixed during Lyapunov certificate updates. 
    However, during policy optimization, where policy parameters are actively updated, sequence-level IS is avoided to prevent prohibitive variance and potential divergence. Following the PPO principle, we utilize only the first-step IS ratio $\rho^i(\phi)$ in (\ref{loss_policy}). This ensures stable optimization while retaining multi-step stability information via the aggregated stability advantage $A_{t,\lambda}$.
\end{remark}

\subsection{Unified Actor-Critic Optimization Framework}
Under the MERL paradigm, the temperature coefficient $\alpha$ in (\ref{J_pi}) is critical for balancing exploration and exploitation. Following the automated entropy adjustment mechanism introduced in \cite{haarnoja2018soft}, we update $\alpha$ by minimizing the following loss function:
\begin{equation}\label{loss_alpha}
    \mathcal{L}_{\text{Ent}}(\alpha) = \frac{1}{Nn} \sum_{i=1}^N \sum_{k=0}^{n-1} \left[ -\alpha \log \pi_\phi(\mathbf{u}_{t+k}^i|\mathbf{x}_{t+k}^i) - \alpha \bar{\mathcal{H}} \right],
\end{equation}
where $\bar{\mathcal{H}}$ denotes the target entropy.

\begin{figure}[!t] 
	\centering
	\includegraphics[width = 3.5in]{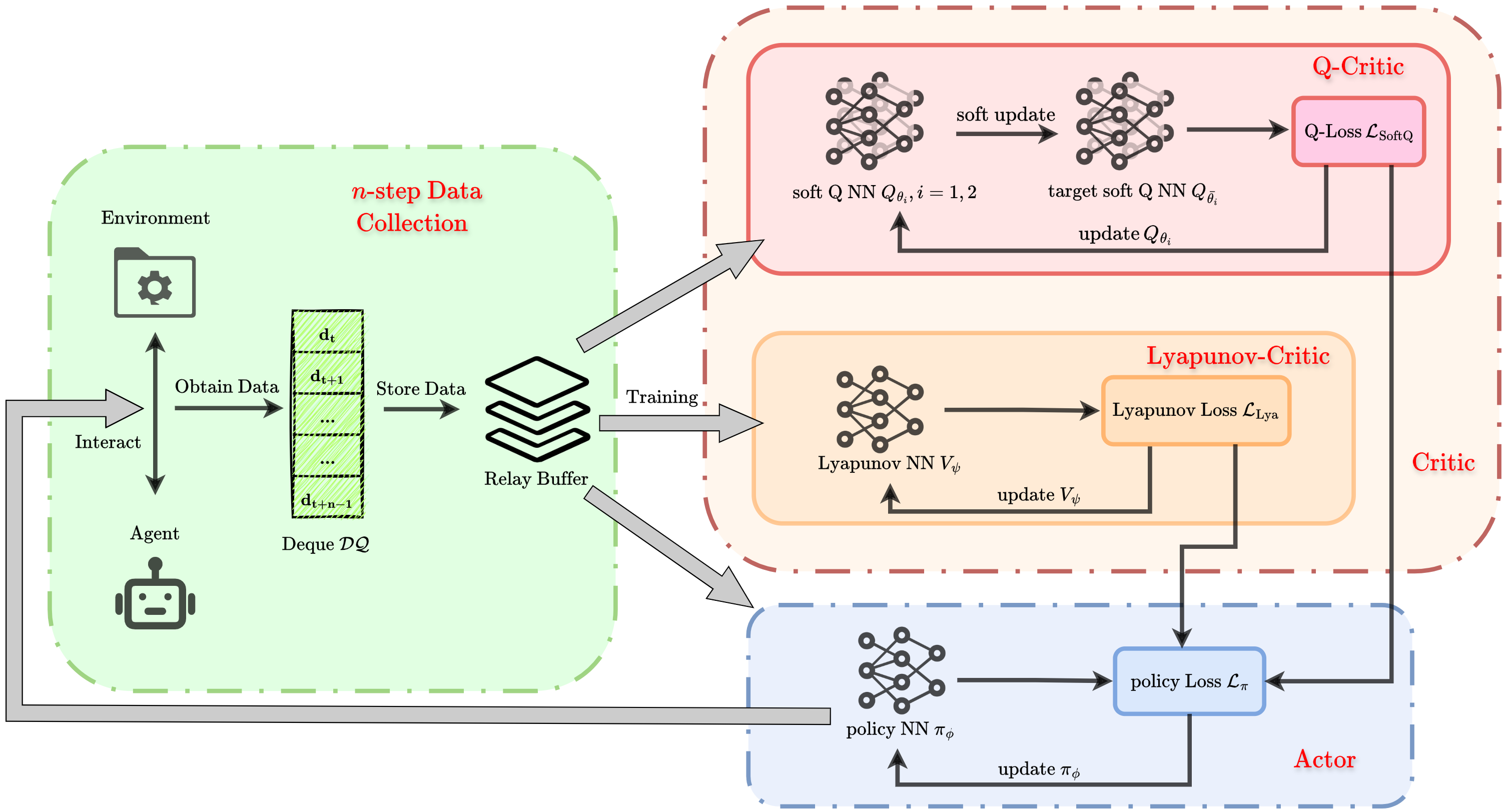}
	\caption{Flowchart of the MSACL algorithm.}
	\label{flow_diagram}
\end{figure}

By integrating the loss functions for the soft Q-network $Q_\theta$, Lyapunov network $V_\psi$, policy network $\pi_\phi$, and temperature coefficient $\alpha$, we iteratively update these parameters within an actor-critic framework. The complete procedure is summarized in Algorithm \ref{msacl} with flowchart illustrated in Fig. \ref{flow_diagram}.

\begin{algorithm}[!ht]
\caption{Multi-Step Actor-Critic Learning with Lyapunov Certificates (MSACL)}
\label{msacl}
\begin{algorithmic}
    \Require {Double soft Q-networks $Q_{\theta_1}, Q_{\theta_2}$, Lyapunov network $V_\psi$, policy network $\pi_\phi$, temperature coefficient $\alpha$}
    \Require {Target soft Q-network parameters $\bar{\theta}_i \leftarrow \theta_i, i=1,2$}
    \Require {Replay buffer $\mathcal{D}$, double-ended queue (deque) $\mathcal{DQ}$ with length $n$, delay frequency $d$}
    \For{each iteration}
        \For{each sampling step}
            \State Select action $\mathbf{u}_t \sim \pi_{\phi}(\cdot|\mathbf{x}_t)$
            \State Obtain reward $r_t$ and observe next state $\mathbf{x}_{t+1}$
            \State Store tuple $\mathbf{d}_t = (\mathbf{x}_t, \mathbf{u}_t, r_t, \pi_{\phi}(\mathbf{u}_t|\mathbf{x}_t), \mathbf{x}_{t+1})$ in $\mathcal{DQ}$
            \If{\text{length}($\mathcal{DQ}$) $= n$}
                \State Store $\{\mathbf{d}_i\}_{i=t}^{t+n-1} = \{\mathbf{d}_{t}, \dots, \mathbf{d}_{t+n-1}\}$ in $\mathcal{D}$
            \EndIf
        \EndFor
        \For{each update step}
            \State Sample $N$ multi-step sequences $\{\mathbf{d}_i\}_{i=t}^{t+n-1}$ from $\mathcal{D}$
            \State Update $V_{\psi}$ via $\psi \leftarrow \psi - \beta_\psi \nabla_\psi \mathcal{L}_{\text{Lya}}(\psi)$
            \State Update $Q_{\theta_i}$ via $\theta_i \leftarrow \theta_i - \beta_\theta \nabla_{\theta_i} \mathcal{L}_{\text{SoftQ}}(\theta_i)$
            \If{$\left(\text{iteration count} \mod d \right) = 0$}
                \State Update $\pi_\phi$ via $\phi \leftarrow \phi - \beta_\phi \nabla_\phi \mathcal{L}_{\pi}(\phi)$
                \State Update $\alpha$ via $\alpha \leftarrow \alpha - \beta_\alpha \nabla_\alpha \mathcal{L}_{\text{Ent}}(\alpha)$
            \EndIf
            \State Update target parameters: $\bar{\theta}_i \leftarrow \tau \theta_i + (1-\tau) \bar{\theta}_i$
        \EndFor  
    \EndFor
\end{algorithmic}
\end{algorithm}

\begin{remark}\label{refinements_remark}
    Algorithm \ref{msacl} incorporates two implementation refinements. First, for the soft Q-loss (\ref{loss_softQ}), we replace $Q_{\theta}$ with $\min\{Q_{\theta_1}, Q_{\theta_2}\}$ to mitigate overestimation bias. Second, we implement delayed updates for both policy $\pi_\phi$ and temperature $\alpha$, performing $d$ consecutive updates within each delayed cycle to ensure sufficient policy improvement. These techniques, adapted from the \verb|cleanrl| framework \cite{huang2022cleanrl}, improve training stability and performance.
\end{remark}

\section{Experiments}\label{sec:experiments}
This section evaluates the MSACL algorithm across six benchmarking environments, covering stabilizing and high-dimensional tracking tasks. We compare our approach against model-free and Lyapunov-based RL baselines using performance and stability metrics. Further analysis includes Lyapunov certificate visualization, robustness assessment under parametric uncertainty and stochastic noise, and sensitivity studies on the multi-step horizon $n$.

\subsection{Benchmarking Environments}

\begin{figure}[!t]
\centering
\subfloat[]{\includegraphics[height=3.0cm, keepaspectratio, width=0.33\columnwidth]{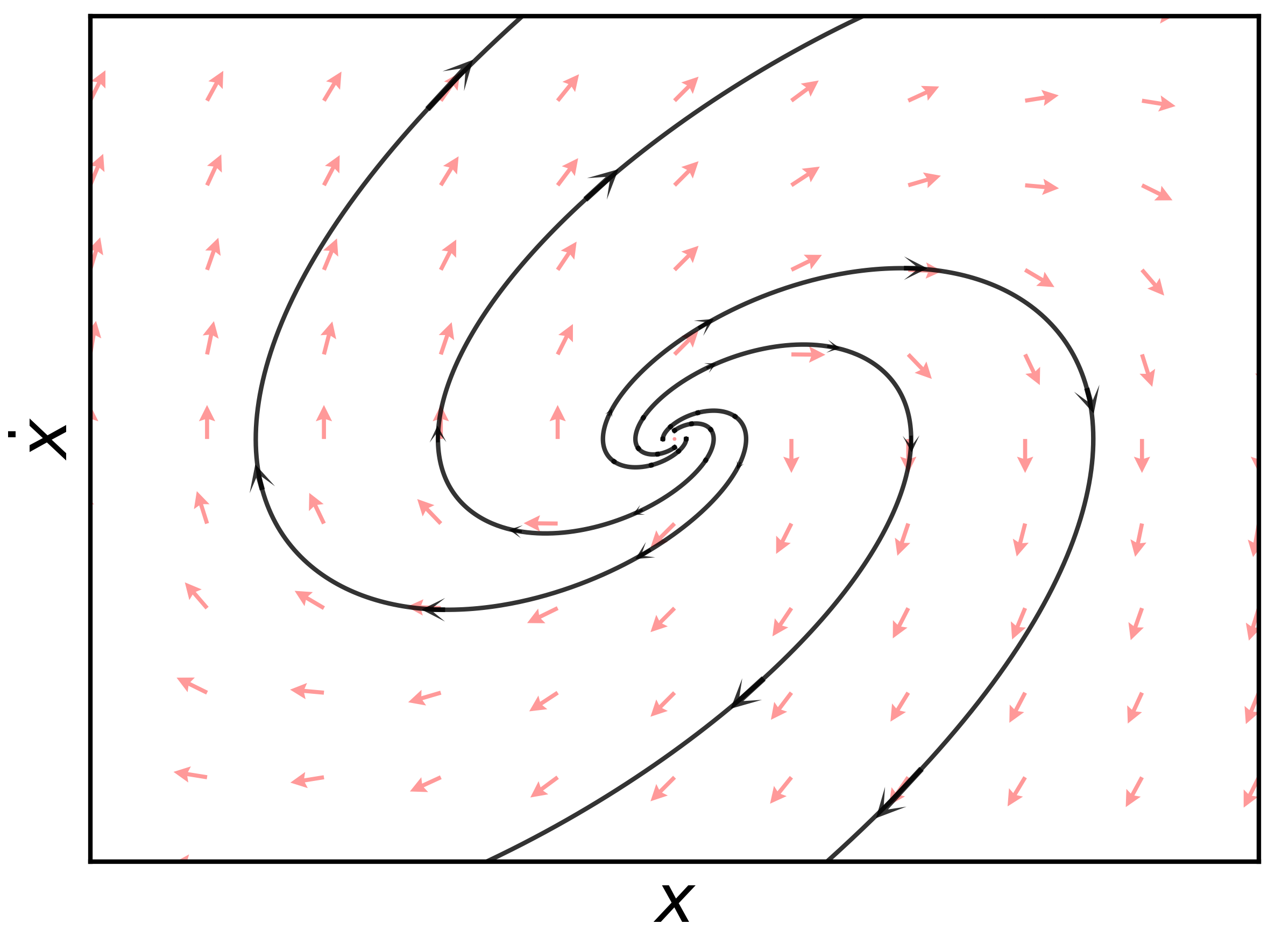}}
\hfill
\subfloat[]{\includegraphics[height=2.2cm, keepaspectratio, width=0.33\columnwidth]{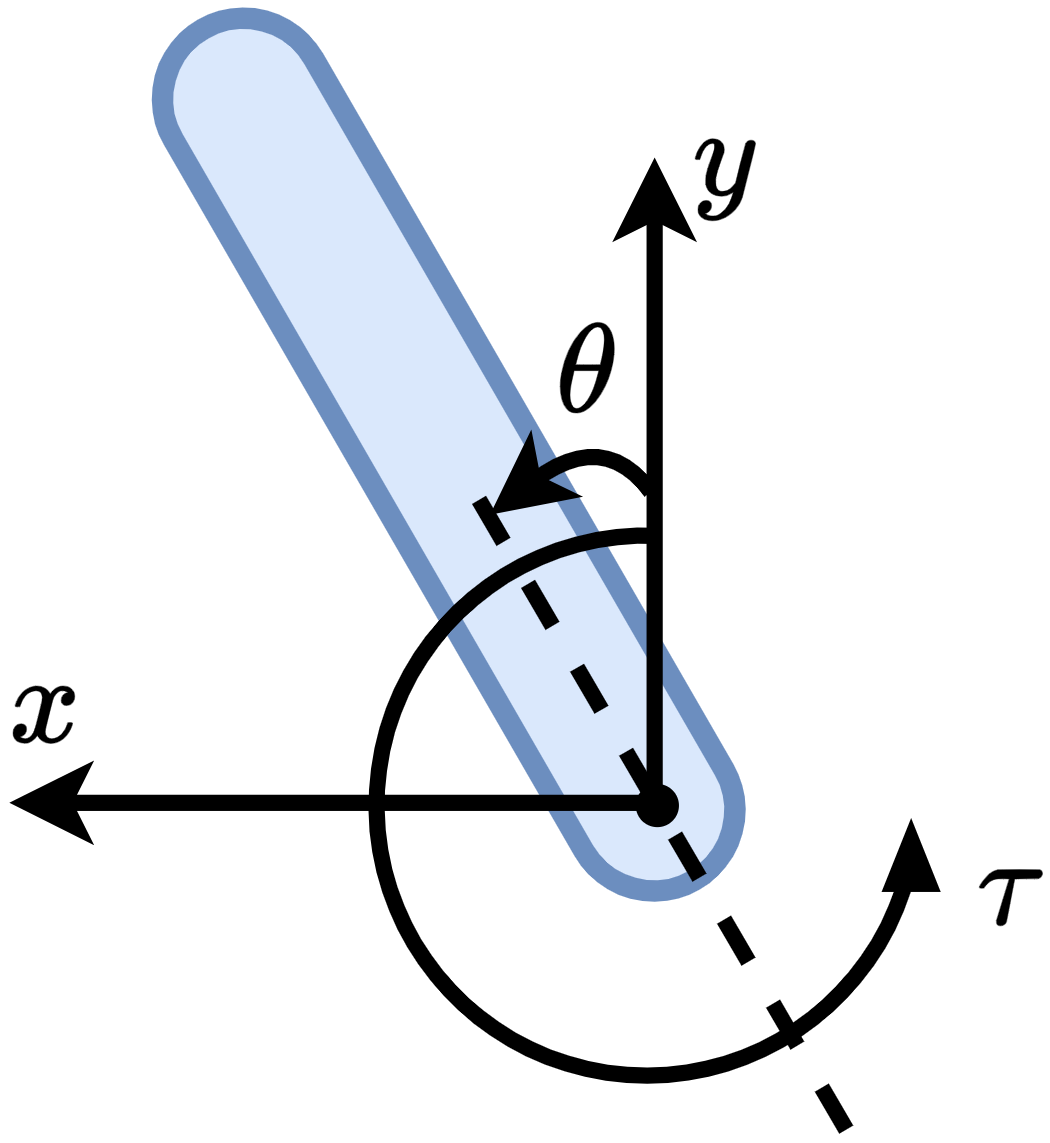}}
\hfill
\subfloat[]{\includegraphics[height=2.5cm, keepaspectratio, width=0.33\columnwidth]{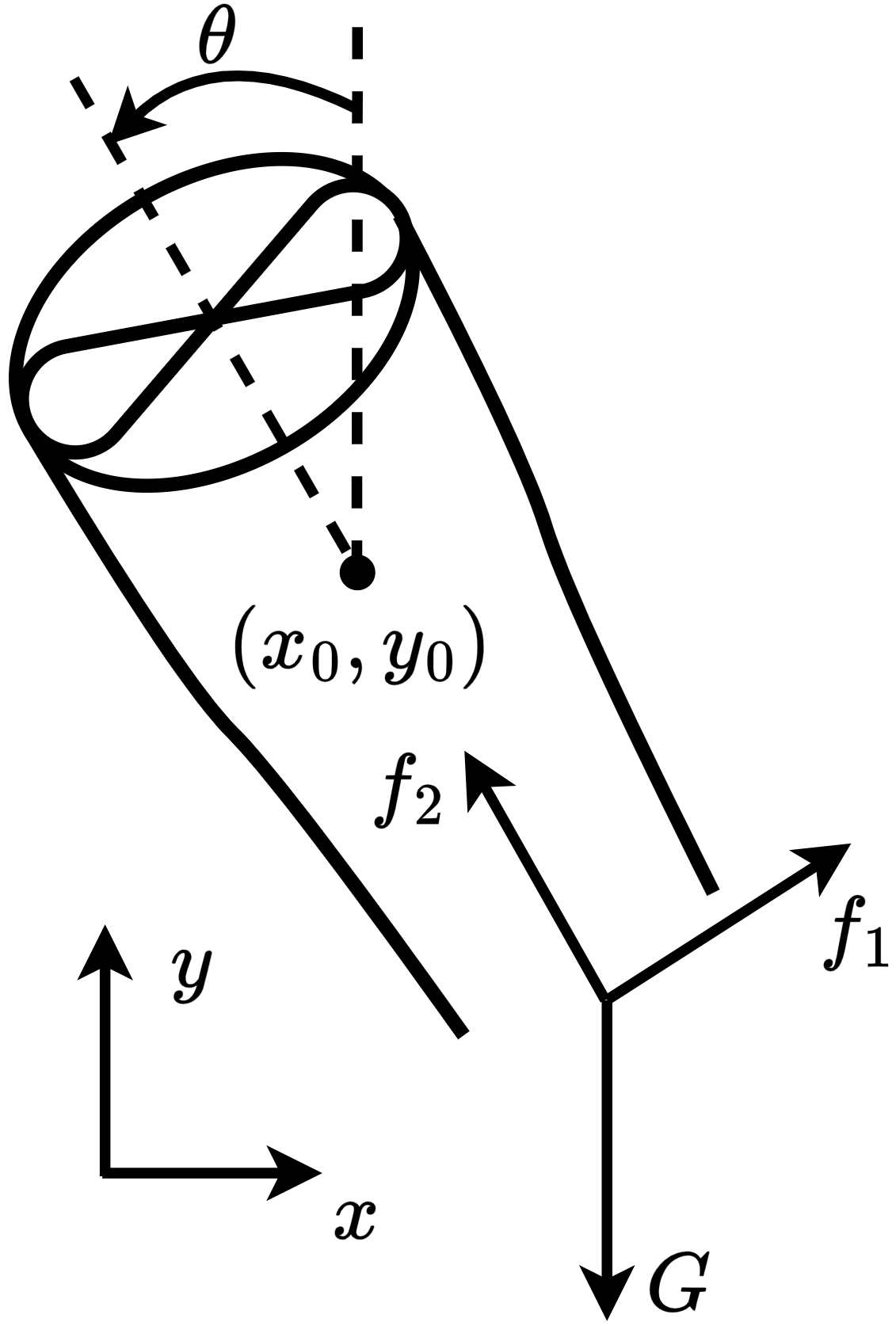}}
\vspace{-5pt}
\subfloat[]{\includegraphics[height=2.5cm, keepaspectratio, width=0.29\columnwidth]{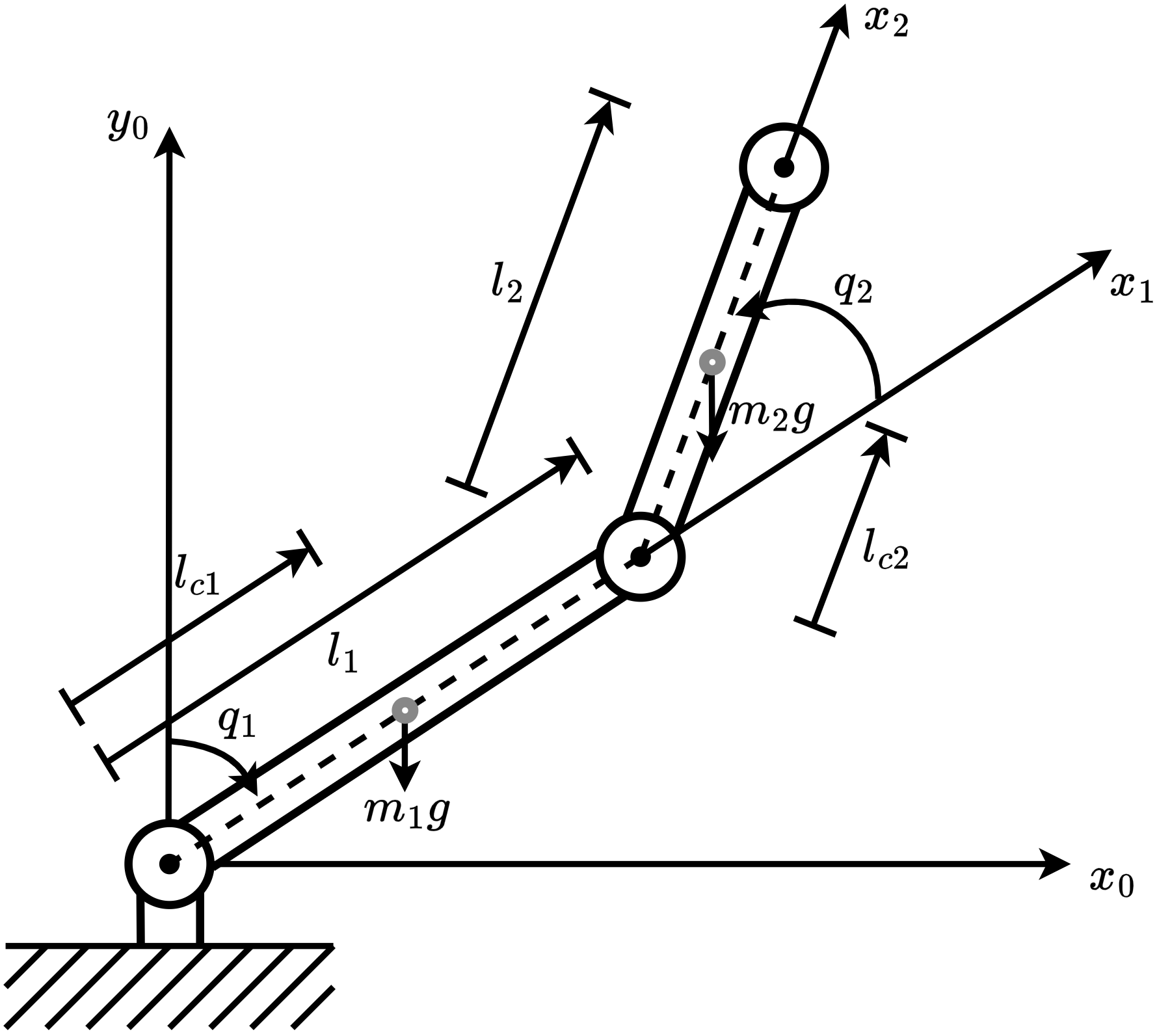}}
\hfill
\subfloat[]{\includegraphics[height=6.0cm, keepaspectratio, width=0.39\columnwidth]{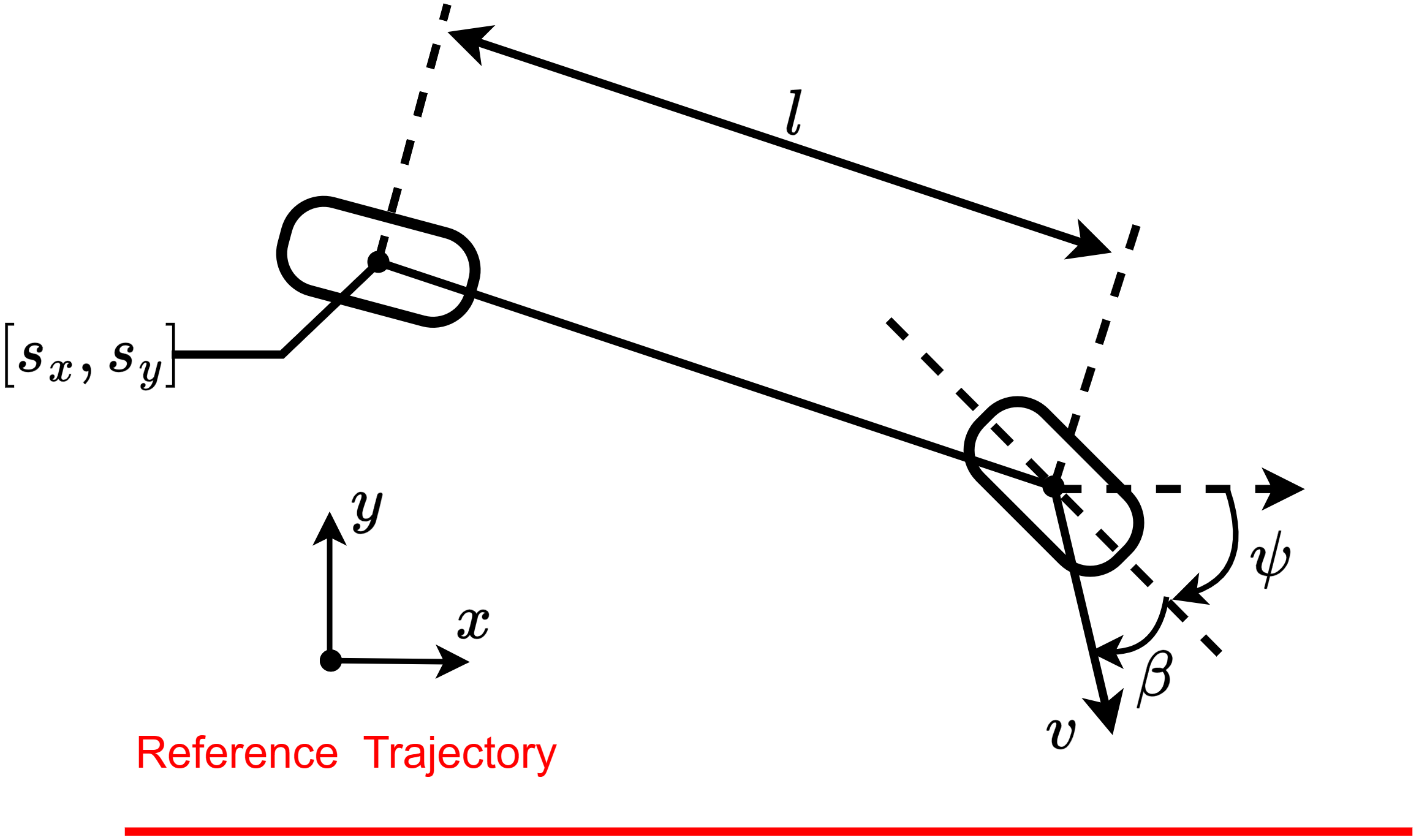}}
\hfill
\subfloat[]{\includegraphics[height=2.5cm, keepaspectratio, width=0.32\columnwidth]{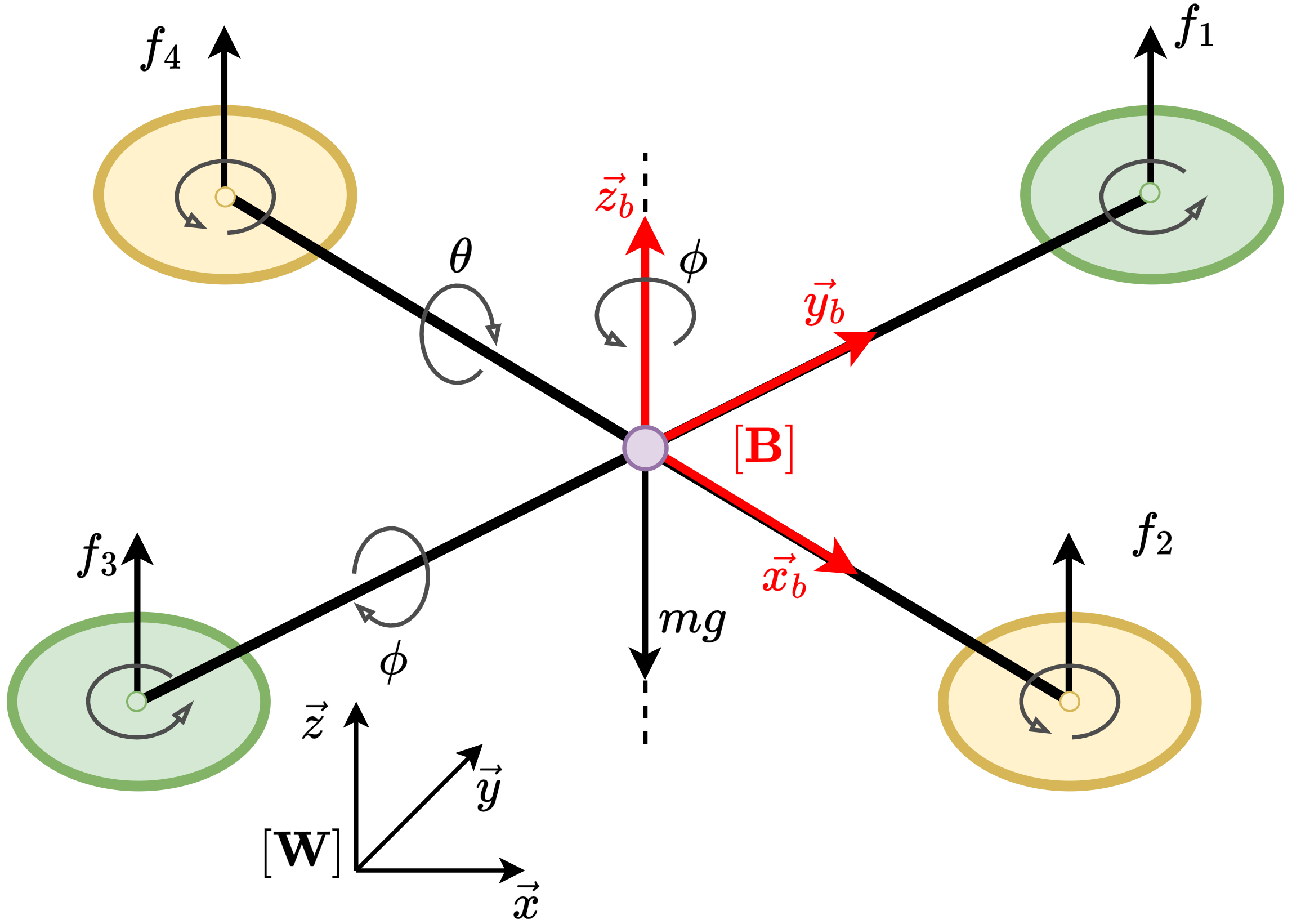}}
\vspace{-5pt}
\caption{Benchmarking environments. (a) VanderPol (Uncontrolled Phase Portrait), (b) Pendulum, (c) Planar DuctedFan, (d) Two-link Planar Robot, (e) SingleCarTracking, (f) QuadrotorTracking.}
\label{benchmark}
\end{figure}

This subsection presents the benchmarking environments for evaluating MSACL and baseline algorithms. We categorize environments by control objective: stabilizing tasks include Controlled VanderPol (VanderPol) \cite{VanderPol1920}, Pendulum \cite{Ogata2010}, Planar DuctedFan (DuctedFan) \cite{yu2001comparison}, and Two-link planar robot (Two-link) \cite{Spong2020}; tracking tasks comprise SingleCarTracking \cite{althoff2017commonroad} and QuadrotorTracking \cite{lee2010geometric}, as illustrated in Figure \ref{benchmark}.

For stabilizing tasks, the goal is to drive the state to the origin $\mathbf{x}_g = \mathbf{0} \in \mathbb{R}^d$; for tracking tasks, the system follows a time-varying reference signal $\mathbf{x}_t^{\text{ref}}$. All tasks use a unified reward structure:
\begin{equation}\label{reward_function}
    r_t = 
    \underbrace{-\left(\mathbf{x}_t^\top \mathbf{Q_r} \mathbf{x}_t + \mathbf{u}_t^\top \mathbf{R_r} \mathbf{u}_t\right)}_{\text{Penalty}}
    + \underbrace{r_{\mathrm{approach}}}_{\text{Encouragement}}
\end{equation}
where the first term penalizes state deviations and control effort using positive definite matrices $\mathbf{Q_r}$ and $\mathbf{R_r}$, minimizing the cumulative cost $\sum_{t=0}^{\infty} \left[\mathbf{x}_t^\top \mathbf{Q_r} \mathbf{x}_t + \mathbf{u}_t^\top \mathbf{R_r} \mathbf{u}_t\right]$ used in optimal control \cite{lewis2012optimal}. For tracking tasks, $\mathbf{x}_t$ is replaced by the tracking error $\mathbf{e}_t = \mathbf{x}_t - \mathbf{x}_t^{\text{ref}}$. The second term $r_{\text{approach}}$ provides sparse encouragement, defined as:
\begin{equation*}
    r_{\mathrm{approach}} = 
    \begin{cases} 
    r_\delta, & \text{if } \|\mathbf{x}_t\|_\infty \leq \delta \\
    0,         & \text{otherwise}
    \end{cases}
\end{equation*}

Specifically, when $\|\mathbf{x}\|_{\infty} < \delta$ (e.g., 0.01), a positive reward $r_\delta$ is provided to incentivize convergence and mitigate steady-state error.
All benchmarks follow the OpenAI Gym framework \cite{brockman2016openai}. Notably, the Pendulum implementation is modified for increased difficulty, featuring expanded state and action spaces, boundary-based termination, and longer episodes, to better differentiate algorithmic performance. Specific details can be found in our code repository.
During interaction, the environment returns a 5-tuple $(\mathbf{x}_t, \mathbf{u}_t, r_t, \pi_\phi(\mathbf{u}_t|\mathbf{x}_t), \mathbf{x}_{t+1})$, with state transitions updated via explicit Euler integration.

\subsection{Baseline Algorithms}
We evaluate MSACL against classic model-free RL baselines (SAC \cite{haarnoja2018soft}, PPO \cite{schulman2017proximal}) and state-of-the-art Lyapunov-based algorithms (LAC \cite{han2020actor}, POLYC \cite{chang2021stabilizing}). SAC and PPO serve as performance baselines for stabilizing and tracking tasks, while LAC and POLYC facilitate stability through Lyapunov constraints. For LAC, the standard reward is replaced with a quadratic cost function defined as:
\begin{equation*}
    c_t = \mathbf{x}_t^\top \mathbf{Q_c} \mathbf{x}_t,
\end{equation*}
where $\mathbf{Q_c}$ is positive definite. Note that in contrast to this state-only penalty formulation, our reward $r_t$ accounts for both state deviations and control effort. By comparing against these baselines, we aim to demonstrate the advantages of MSACL in terms of performance and stability metrics introduced in the next subsection.

For fair comparison, all algorithms are implemented in the \verb|GOPS| framework \cite{wang2023gops}. Actor and critic networks utilize MLPs with two 256-unit hidden layers and ReLU activations. All actors employ stochastic diagonal Gaussian policies. Critic architectures vary: LAC uses only a Lyapunov network, while POLYC incorporates both value and Lyapunov networks. SAC and LAC also implement clipped double Q-learning and delayed policy updates according to Remark \ref{refinements_remark}. Optimization is performed via the Adam optimizer \cite{kingma2014adam}, with core hyperparameters listed in Table \ref{para_table}.

\vspace{-5pt}
\begin{table}[!ht]\label{para_table}
    \centering
    \caption{Hyperparameters in the algorithms}
    \begin{tabular}{lc}
    \toprule
    Hyperparameters & Value \\
    \midrule
    \textit{Shared} & \\
    \quad Actor learning rate & $3\text{e}-4$ \\
    \quad Critic learning rate & $1\text{e}-3$ \\
    \quad Discount factor ($\gamma$) & 0.99 \\
    \quad Policy delay update interval ($d$) & 2 \\
    \quad Target smoothing coefficient ($\tau$) & 0.05 \\
    \quad Optimizer & Adam (default $\beta_1$ and $\beta_2$) \\
    \hline
    \textit{Maximum-entropy framework} & \\
    \quad Learning rate of $\alpha$ & $1\text{e}-3$ \\
    \quad Target entropy ($\overline{\mathcal{H}}$) & $\overline{\mathcal{H}} = -\text{dim}(\mathcal{U}) = -m$ \\
    \hline
    \textit{Off-policy} & \\
    \quad Replay buffer warm size & $5 \times 10^3$ \\
    \quad Replay buffer size & $1 \times 10^6$ \\
    \quad Samples collected per iteration & 20 \\
    \quad Replay batch size & 256 \\
    \hline
    \textit{On-policy} & \\
    \quad Sample batch size & 1600 \\
    \quad Replay batch size & 1600 \\
    \quad GAE factor & 0.95 \\
    \hline
    \textit{MSACL} & \\
    \quad $\alpha_1$  & 1 \\
    \quad $\alpha_2$  & 2 \\
    \quad $\alpha_3$  & 0.15 (0.125 is also acceptable) \\
    \quad $\omega_{\text{Bnd}}$ in (\ref{loss_lya}) & 1 \\
    \quad $\omega_{\text{Stab}}$ in (\ref{loss_lya}) & 10 \\
    \quad $\varepsilon$ in (\ref{loss_policy})  & 0.1 \\
    \bottomrule
    \end{tabular}
\end{table}
\vspace{-5pt}

\subsection{Performance Evaluation and Stability Analysis}\label{training_results}
\begin{figure*}[!b]
    \centering
    \subfloat[VanderPol]{\includegraphics[width=0.49\textwidth]{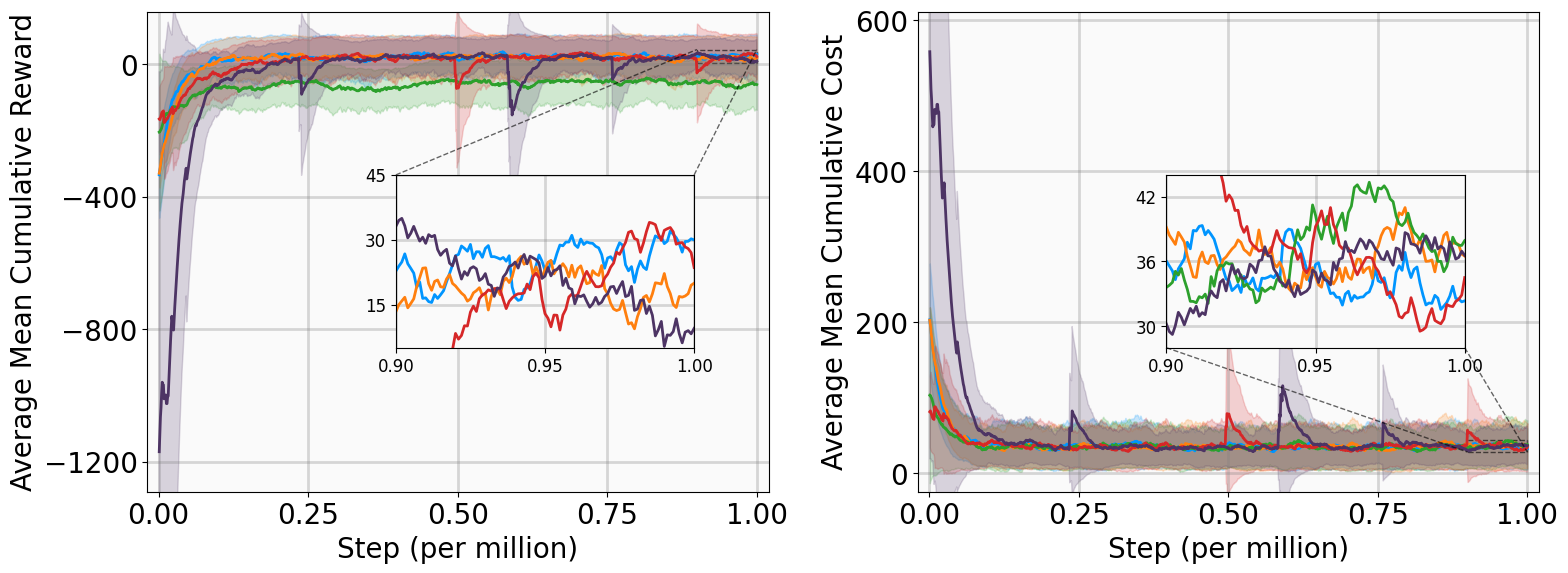}}\hfill
    \subfloat[Pendulum]{\includegraphics[width=0.49\textwidth]{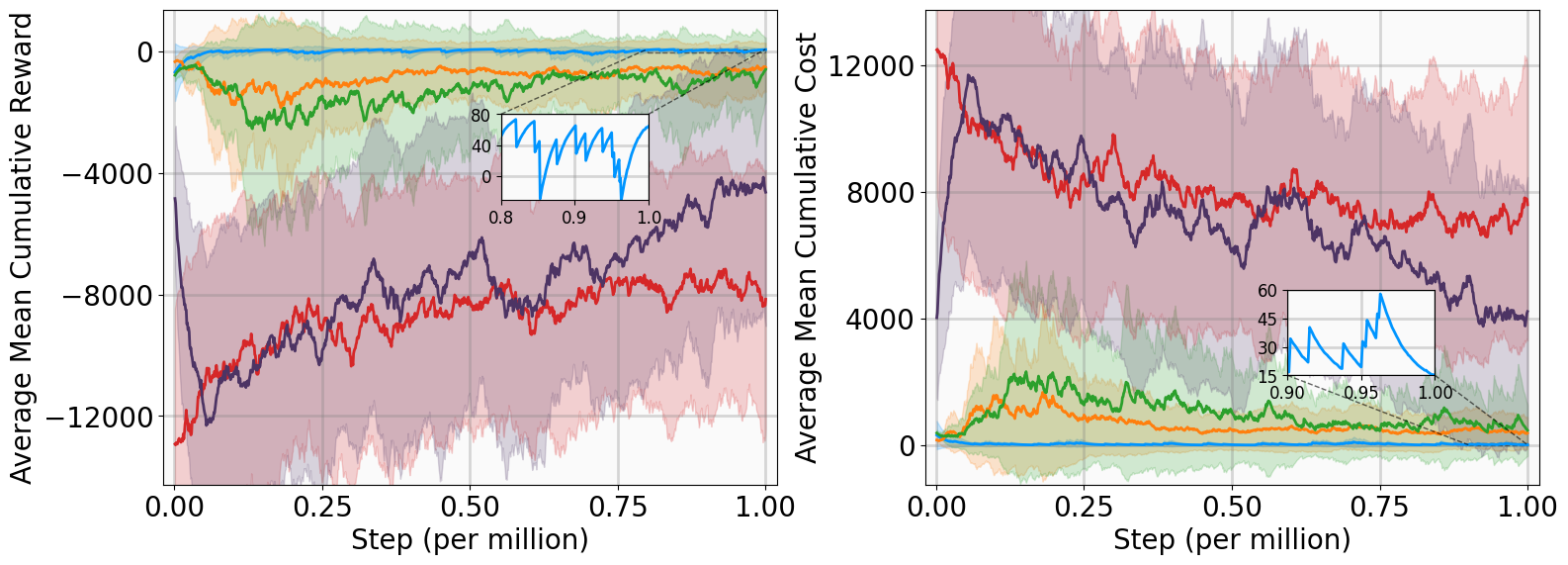}}
    \vspace{-5pt}
    \subfloat[DuctedFan]{\includegraphics[width=0.49\textwidth]{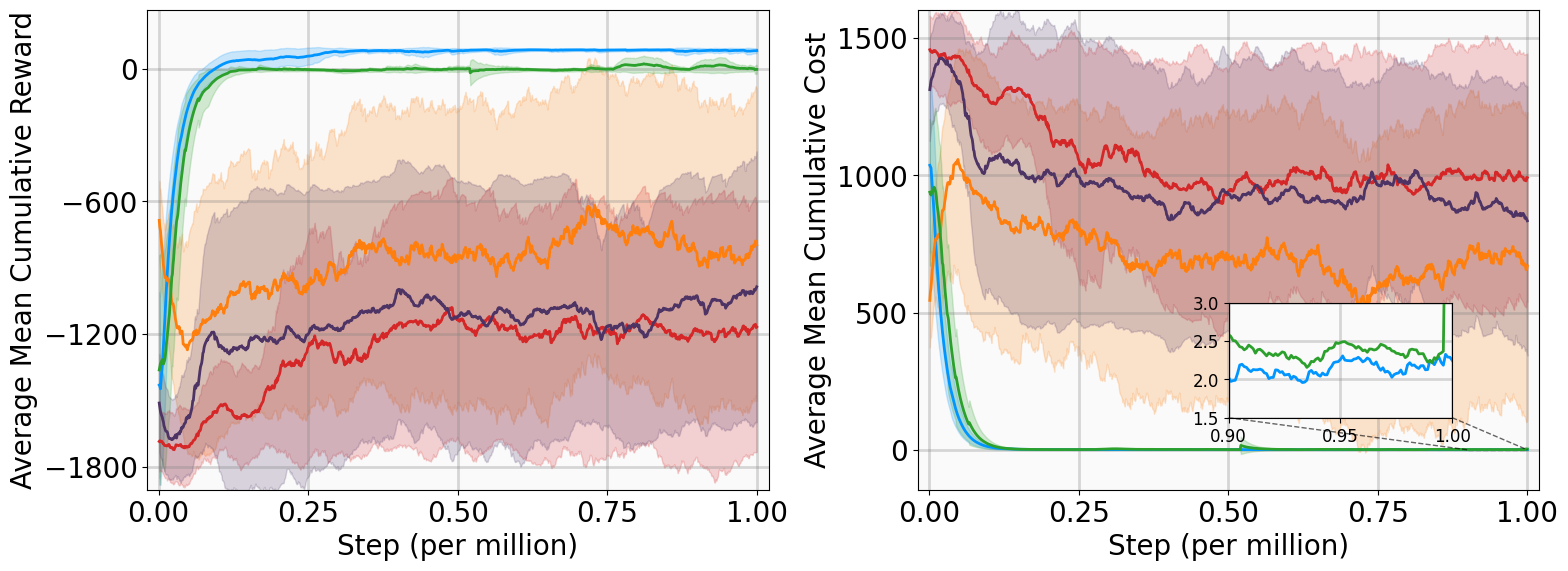}}\hfill
    \subfloat[Two-link]{\includegraphics[width=0.49\textwidth]{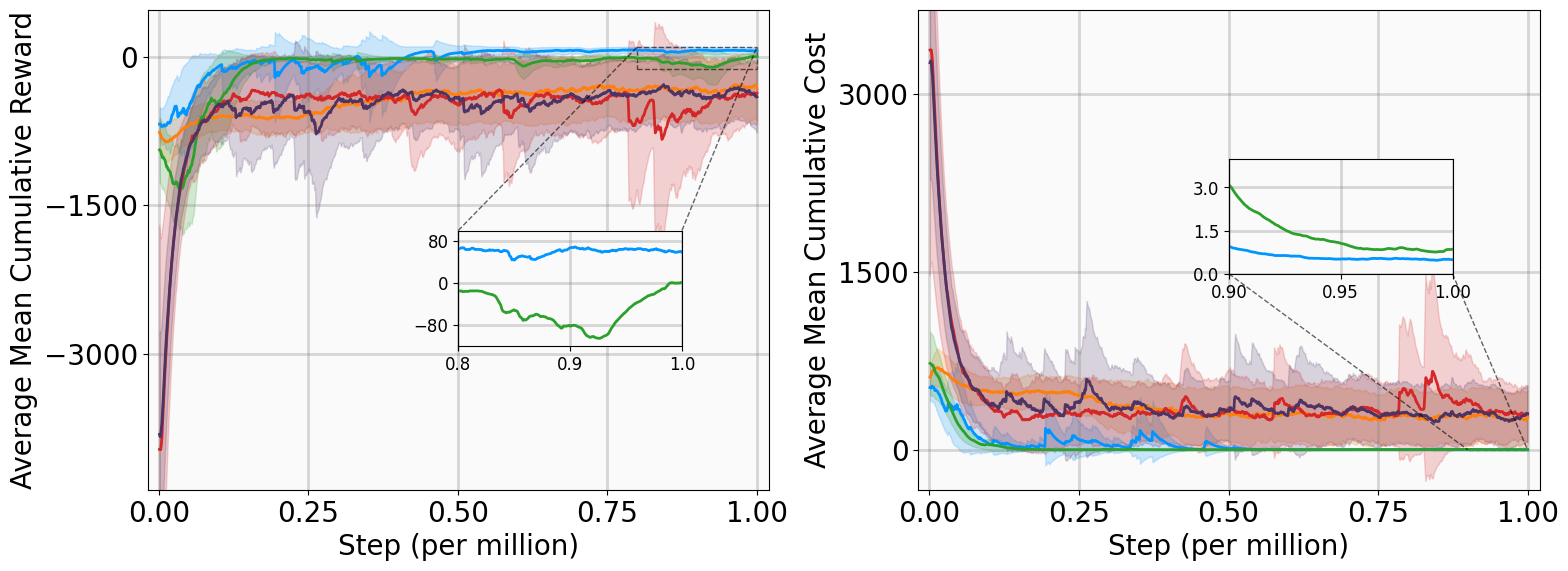}}
    \vspace{-5pt} 
    \subfloat[SingleCarTracking]{\includegraphics[width=0.49\textwidth]{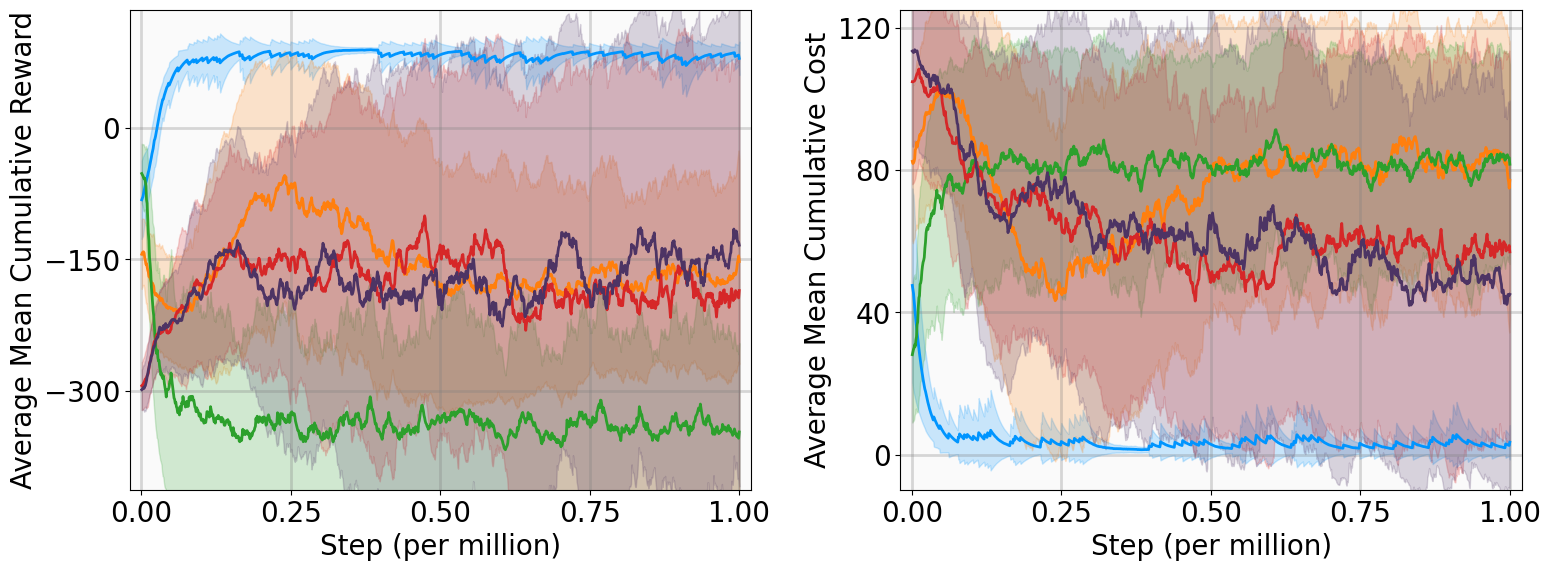}}\hfill
    \subfloat[QuadrotorTracking]{\includegraphics[width=0.49\textwidth]{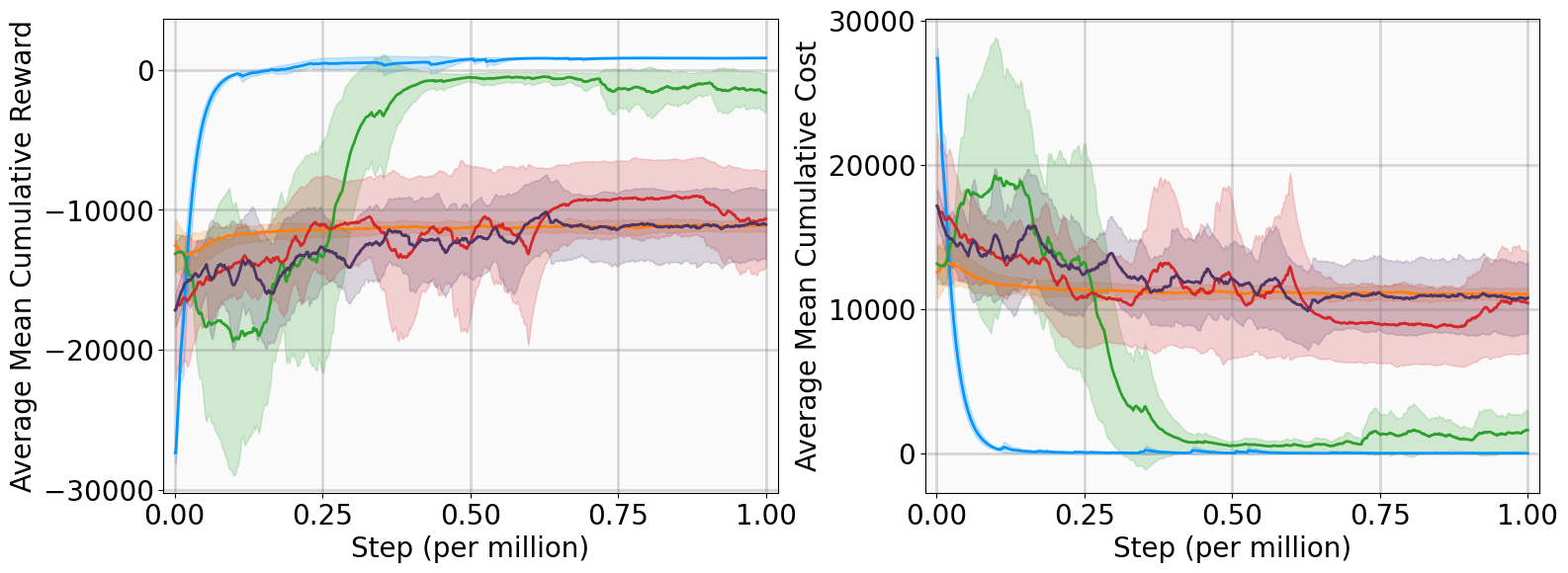}}
    \vspace{6pt} 
    \centering
    \small
    \legenditem{colorMSACL}{MSACL (Ours)} \quad \,\,\,\,
    \legenditem{colorSAC}{SAC \cite{haarnoja2018soft}} \quad \,\,\,\,
    \legenditem{colorLAC}{LAC \cite{han2020actor}} \quad \,\,\,\,
    \legenditem{colorPPO}{PPO \cite{schulman2017proximal}} \quad \,\,\,\,
    \legenditem{colorPOLYC}{POLYC \cite{chang2021stabilizing}}
    \caption{Training performance across six benchmarks. The results show the average MCR and MCC   over five independent runs. Solid lines and shaded regions represent the mean and one standard deviation, respectively. The horizontal axis indicates the number of training steps.}
    \label{training_reward_cost}
\end{figure*}

This subsection presents training outcomes, stability metrics, and Lyapunov certificate visualizations. Each algorithm is trained across five random seeds, with training results illustrated in Fig. \ref{training_reward_cost} using the following performance metrics:
\begin{itemize}
    \item \textbf{AMCR (Average Mean Cumulative Reward):} The average of the Mean Cumulative Reward (MCR) computed across five independent training trials. MCR represents the total undiscounted reward accumulated per episode, normalized by the episode length to measure per-step performance.
    \item \textbf{AMCC (Average Mean Cumulative Cost):} The average of the Mean Cumulative Cost (MCC) computed across five independent training trials. MCC represents the total undiscounted cost accumulated per episode, normalized by the episode length to quantify the average per-step magnitude of state deviations.
\end{itemize}

Training curves show MSACL outperforms or at least matches all baselines across six benchmarks. A positive AMCR indicates that the system state successfully enters the target $\delta$-neighborhood, while higher reward values imply faster convergence as the agent spends more time receiving $r_\delta$.

For comparison, we select the \textit{optimal policy} for each algorithm, defined as the model that achieved the highest MCR among five training runs. These optimal policies are evaluated from 100 fixed initial states, with stability and convergence are assessed across four target radii ($0.2, 0.1, 0.05, 0.01$) using the following stability metrics:
\begin{itemize}
    \item \textbf{Reach Rate (RR):} Percentage of trajectories that enter the specified radius.
    \item \textbf{Average Reach Step (ARS):} Mean steps that required for the state to first enter the radius.
    \item \textbf{Average Hold Step (AHS):} Mean steps that the state remains within the radius after the initial entry.
\end{itemize}
The evaluation results are summarized in Table \ref{algorithm_eval_metrics}.

\begin{table*}[!ht] 
    \centering
    \begin{threeparttable}
    \caption{Experimental Results of Baseline Algorithms Across Benchmarking Environments}
    \label{algorithm_eval_metrics}
    \begin{tabular}{cc cc *{4}{wc{1.7cm}}}
        \toprule
        \multirow{2}{*}{\textbf{Benchmark}} & \multirow{2}{*}{\textbf{Algorithm}} & \multirow{2}{*}{\textbf{AMCR}} & \multirow{2}{*}{\textbf{AMCC}} & \multicolumn{4}{c}{\textbf{RR\,/\,ARS\,/\,AHS at different radii}} \\
        \cmidrule(lr){5-8}
        & & & & 0.2 & 0.1 & 0.05 & 0.01 \\
        \midrule
        
        \multirow{5}{*}{VanderPol}
        & MSACL & \textbf{33.2 $\pm$ 64.5} & 31.0 $\pm$ 29.4 & 1.0\,/\,57.6\,/\,943.2 & \textbf{1.0\,/\,58.4\,/\,942.5} & \textbf{1.0\,/\,58.8\,/\,942.1} & \textbf{1.0\,/\,59.0\,/\,941.9} \\
        & SAC   & 32.3 $\pm$ 64.4 & 31.1 $\pm$ 29.5 & \textbf{1.0\,/\,56.2\,/\,944.6} & 1.0\,/\,58.9\,/\,942.0 & 1.0\,/\,61.0\,/\,939.9 & 1.0\,/\,67.6\,/\,933.3 \\
        & LAC   & 21.2 $\pm$ 65.1 & 31.1 $\pm$ 29.7 & 1.0\,/\,73.1\,/\,927.8 & 1.0\,/\,93.4\,/\,907.5 & 1.0\,/\,112.0\,/\,888.9 & 1.0\,/\,168.6\,/\,832.3 \\
        & PPO   & 27.1 $\pm$ 64.1 & \textbf{30.8 $\pm$ 29.4} & 1.0\,/\,65.6\,/\,935.3 & 1.0\,/\,77.8\,/\,923.1 & 1.0\,/\,87.6\,/\,913.4 & 1.0\,/\,110.8\,/\,890.1 \\
        & POLYC & 32.3 $\pm$ 64.5 & \textbf{30.8 $\pm$ 29.4} & 1.0\,/\,59.3\,/\,941.6 & 1.0\,/\,61.6\,/\,939.3 & 1.0\,/\,63.2\,/\,937.8 & 1.0\,/\,66.4\,/\,934.5 \\
        \cmidrule(lr){1-8}
        
        \multirow{5}{*}{Pendulum} 
        & MSACL & \textbf{35.4 $\pm$ 360.1} & \textbf{30.8 $\pm$ 171.7} & \textbf{0.98\,/\,35.3\,/\,965.6} & \textbf{0.98\,/\,40.3\,/\,960.6} & \textbf{0.98\,/\,41.2\,/\,959.7} & \textbf{0.98\,/\,42.5\,/\,958.4} \\
        & SAC   & -544.4 $\pm$ 1073.4 & 382.1 $\pm$ 640.0 & 0.74\,/\,34.0\,/\,966.8 & 0.74\,/\,36.0\,/\,964.9 & 0.74\,/\,37.0\,/\,963.9 & 0.74\,/\,41.6\,/\,959.3 \\
        & LAC   & -392.6 $\pm$ 899.9 & 269.2 $\pm$ 503.4 & 0.78\,/\,37.6\,/\,963.3 & 0.78\,/\,51.0\,/\,950.0 & 0.78\,/\,64.6\,/\,936.4 & 0.78\,/\,111.6\,/\,889.4 \\
        & PPO   & -6443.4 $\pm$ 4827.7 & 5843.8 $\pm$ 4912.2 & 0.0\,/\,--\,/\,-- & 0.0\,/\,--\,/\,-- & 0.0\,/\,--\,/\,-- & 0.0\,/\,--\,/\,-- \\
        & POLYC & -2576.9 $\pm$ 4404.2 & 2379.5 $\pm$ 3986.4 & 0.69\,/\,27.8\,/\,972.4 & 0.69\,/\,33.6\,/\,967.2 & 0.69\,/\,37.8\,/\,963.1 & 0.69\,/\,45.4\,/\,955.5 \\
        \cmidrule(lr){1-8}
        
        \multirow{5}{*}{Ducted Fan} 
        & MSACL & \textbf{88.6 $\pm$ 3.5} & 2.1 $\pm$ 1.3 & \textbf{1.0\,/\,35.3\,/\,965.3} & \textbf{1.0\,/\,52.7\,/\,948.0} & \textbf{1.0\,/\,59.6\,/\,941.3} & \textbf{1.0\,/\,68.9\,/\,932.0} \\
        & SAC   & -12.6 $\pm$ 433.8 & 81.1 $\pm$ 347.0 & 0.95\,/\,32.3\,/\,968.1 & 0.95\,/\,46.3\,/\,954.6 & 0.95\,/\,55.4\,/\,945.5 & 0.95\,/\,88.6\,/\,910.9 \\
        & LAC   & 84.2 $\pm$ 4.0 & \textbf{2.0 $\pm$ 1.3} & 1.0\,/\,40.6\,/\,959.8 & 1.0\,/\,58.7\,/\,941.8 & 1.0\,/\,70.1\,/\,930.0 & 1.0\,/\,102.0\,/\,898.8 \\
        & PPO   & -8.5 $\pm$ 372.7 & 73.0 $\pm$ 285.4 & 0.94\,/\,40.6\,/\,958.9 & 0.94\,/\,59.4\,/\,941.3 & 0.94\,/\,76.6\,/\,924.1 & 0.94\,/\,113.4\,/\,887.4 \\
        & POLYC & 82.3 $\pm$ 4.3 & 2.4 $\pm$ 1.4 & 1.0\,/\,44.8\,/\,955.3 & 1.0\,/\,68.7\,/\,932.0 & 1.0\,/\,87.4\,/\,913.1 & 1.0\,/\,130.5\,/\,870.4 \\
        \cmidrule(lr){1-8}
        
        \multirow{5}{*}{Two-link} 
        & MSACL & \textbf{89.0 $\pm$ 7.7} & 0.4 $\pm$ 0.3 & 1.0\,/\,14.5\,/\,986.3 & \textbf{1.0\,/\,22.0\,/\,978.9} & \textbf{1.0\,/\,27.9\,/\,973.0} & \textbf{1.0\,/\,41.5\,/\,957.2} \\
        & SAC   & -159.6 $\pm$ 362.5 & 179.4 $\pm$ 257.3 & 0.67\,/\,12.4\,/\,988.4 & 0.67\,/\,18.4\,/\,982.4 & 0.67\,/\,24.3\,/\,976.6 & 0.67\,/\,36.9\,/\,964.0 \\
        & LAC   & 82.5 $\pm$ 8.2 & \textbf{0.3 $\pm$ 0.2} & \textbf{1.0\,/\,14.4\,/\,986.5} & 1.0\,/\,28.7\,/\,972.2 & 1.0\,/\,45.9\,/\,955.0 & 1.0\,/\,92.6\,/\,908.3 \\
        & PPO   & -676.9 $\pm$ 863.7 & 472.7 $\pm$ 566.2 & 0.48\,/\,12.4\,/\,988.5 & 0.48\,/\,18.1\,/\,982.6 & 0.48\,/\,24.2\,/\,976.7 & 0.48\,/\,46.9\,/\,954.0 \\
        & POLYC & -172.0 $\pm$ 335.2 & 190.6 $\pm$ 246.6 & 0.60\,/\,14.8\,/\,939.0 & 0.60\,/\,22.8\,/\,978.1 & 0.60\,/\,28.9\,/\,972.0 & 0.60\,/\,40.2\,/\,960.7 \\
        \cmidrule(lr){1-8}
        
        \multirow{5}{*}{SingleCarTracking} 
        & MSACL & \textbf{84.7 $\pm$ 32.7} & \textbf{2.9 $\pm$ 14.1} & \textbf{0.99\,/\,37.6\,/\,961.6} & \textbf{0.99\,/\,55.3\,/\,940.7} & \textbf{0.99\,/\,76.5\,/\,924.1} & \textbf{0.99\,/\,96.4\,/\,904.5} \\
        & SAC   & 10.3 $\pm$ 164.9 & 21.9 $\pm$ 50.3 & 0.85\,/\,36.5\,/\,962.9 & 0.85\,/\,54.3\,/\,946.6 & 0.85\,/\,81.4\,/\,919.5 & 0.85\,/\,193.3\,/\,807.6 \\
        & LAC   & -350.5 $\pm$ 196.8 & 94.5 $\pm$ 47.4 & 0.17\,/\,14.2\,/\,986.7 & 0.17\,/\,51.2\,/\,949.6 & 0.17\,/\,107.5\,/\,891.0 & 0.17\,/\,221.1\,/\,763.4 \\
        & PPO   & 10.2 $\pm$ 188.6 & 21.5 $\pm$ 49.2 & 0.85\,/\,38.6\,/\,961.8 & 0.85\,/\,61.2\,/\,937.6 & 0.85\,/\,80.6\,/\,919.1 & 0.85\,/\,102.7\,/\,894.4 \\
        & POLYC & 40.1 $\pm$ 152.8 & 14.7 $\pm$ 39.9 & 0.90\,/\,34.9\,/\,962.0 & 0.90\,/\,58.0\,/\,939.8 & 0.90\,/\,67.6\,/\,932.3 & 0.90\,/\,81.8\,/\,912.3 \\
        \cmidrule(lr){1-8}
        
        \multirow{5}{*}{QuadrotorTracking} 
        & MSACL & \textbf{838.8 $\pm$ 1.5} & \textbf{14.5 $\pm$ 2.1} & \textbf{1.0\,/\,33.1\,/\,967.8} & \textbf{1.0\,/\,34.0\,/\,966.9} & \textbf{1.0\,/\,36.9\,/\,964.0} & \textbf{1.0\,/\,49.7\,/\,619.8} \\
        & SAC   & -17263.8 $\pm$ 1380.0 & 17246.4 $\pm$ 1379.1 & 0.0\,/\,--\,/\,-- & 0.0\,/\,--\,/\,-- & 0.0\,/\,--\,/\,-- & 0.0\,/\,--\,/\,-- \\
        & LAC   & -51.1 $\pm$ 1.2 & 32.7 $\pm$ 1.1 & 1.0\,/\,471.9\,/\,20.6 & 0.0\,/\,--\,/\,-- & 0.0\,/\,--\,/\,-- & 0.0\,/\,--\,/\,-- \\
        & PPO   & -12836.6 $\pm$ 4042.3 & 12571.3 $\pm$ 4022.4 & 0.0\,/\,--\,/\,-- & 0.0\,/\,--\,/\,-- & 0.0\,/\,--\,/\,-- & 0.0\,/\,--\,/\,-- \\
        & POLYC & -17528.1 $\pm$ 2162.6 & 17318.7 $\pm$ 2155.6 & 0.0\,/\,--\,/\,-- & 0.0\,/\,--\,/\,-- & 0.0\,/\,--\,/\,-- & 0.0\,/\,--\,/\,-- \\
        \bottomrule
    \end{tabular}
    \begin{tablenotes}[flushleft]
        \footnotesize
        \item \textbf{Note:} AMCR and AMCC values represent the mean performance of the optimal policy (highest evaluation MCR) averaged over 100 independent trials, with $\pm$ denoting one standard deviation. Stability metrics (RR, ARS, and AHS) are evaluated across four radii; -- indicates the radius was not reached. Bold entries highlight superior results: maximum AMCR, minimum AMCC, and highest RR. For stability metrics, the highest RR is prioritized for bolding, with the shortest ARS and longest AHS bolded for identical RR values.
    \end{tablenotes}
    \end{threeparttable}
\end{table*}

Evaluation results demonstrate MSACL's superiority. Specifically, MSACL consistently achieves the highest AMCR across all six benchmarks. Regarding AMCC, MSACL attains the minimum cost among all evaluated methods in the Pendulum, SingleCarTracking, and QuadrotorTracking tasks, while remaining competitive in others. Furthermore, MSACL maintains the highest RR across all radii. Notably, in the high-dimensional QuadrotorTracking task, MSACL is the only approach to achieve a 100\% reach rate at the $0.01$ precision level, whereas baselines fail. These results demonstrate the speed and precision of MSACL in complex control systems.

\begin{figure*}[!ht]
    \centering
    \subfloat[]{\includegraphics[width=0.19\textwidth]{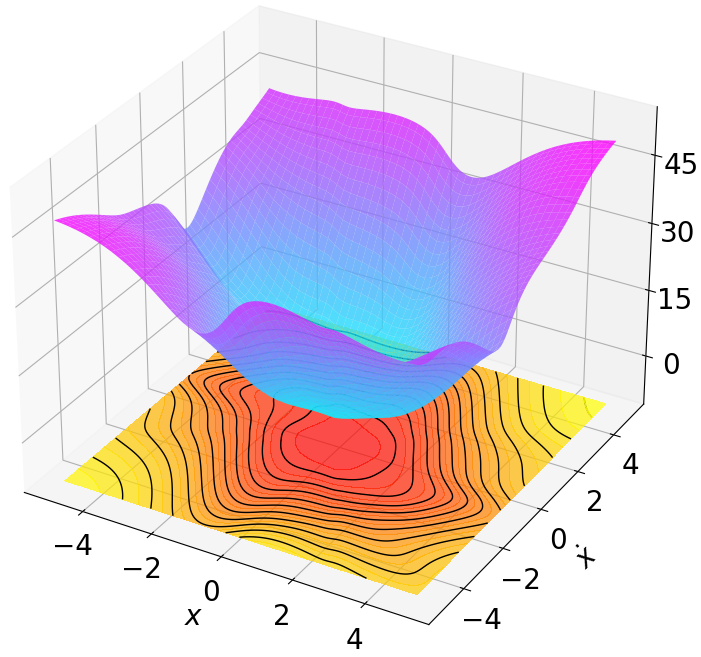}}\hfill
    \subfloat[]{\includegraphics[width=0.19\textwidth]{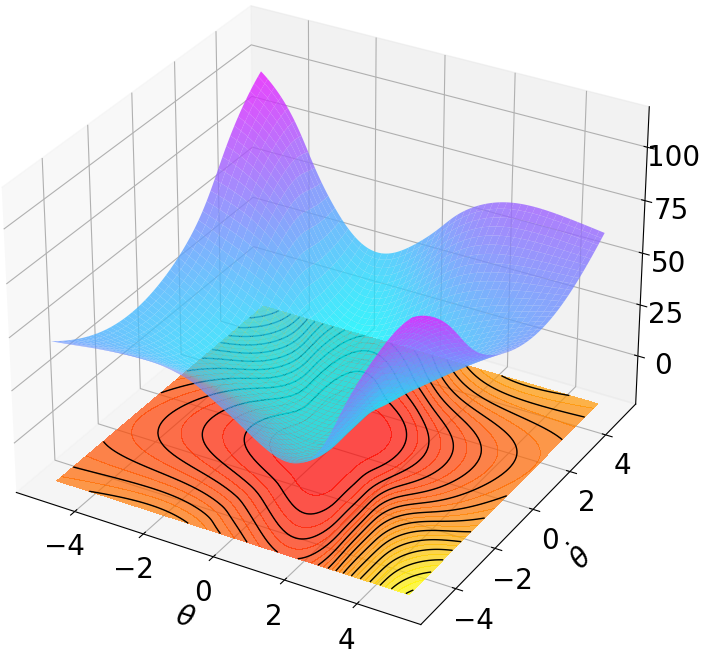}}\hfill
    \subfloat[]{\includegraphics[width=0.19\textwidth]{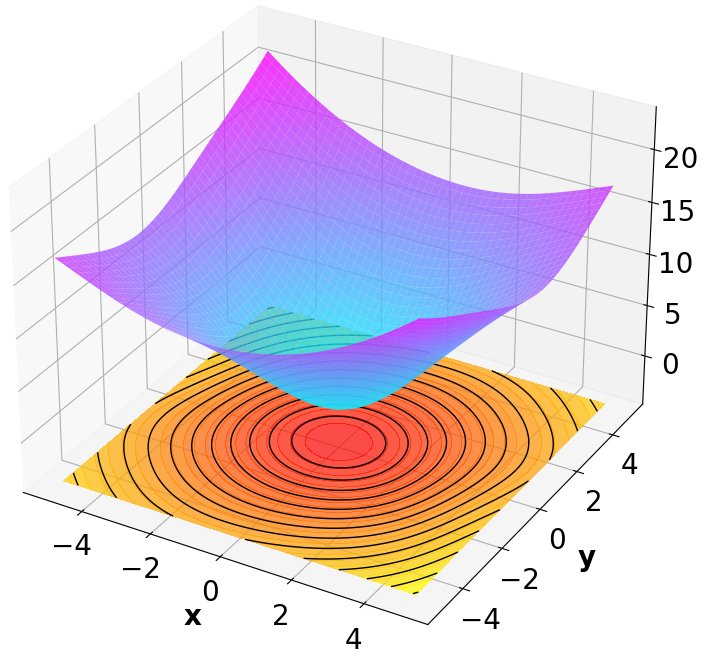}}\hfill
    \subfloat[]{\includegraphics[width=0.19\textwidth]{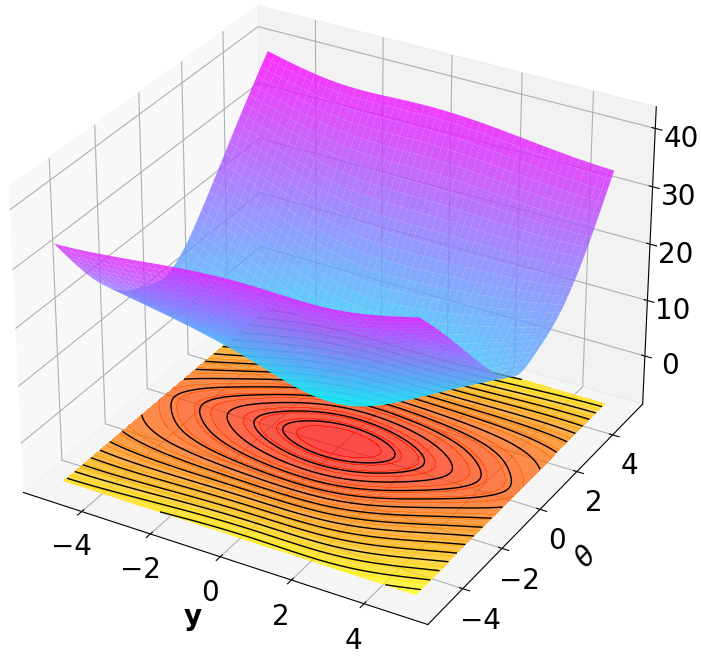}}\hfill
    \subfloat[]{\includegraphics[width=0.19\textwidth]{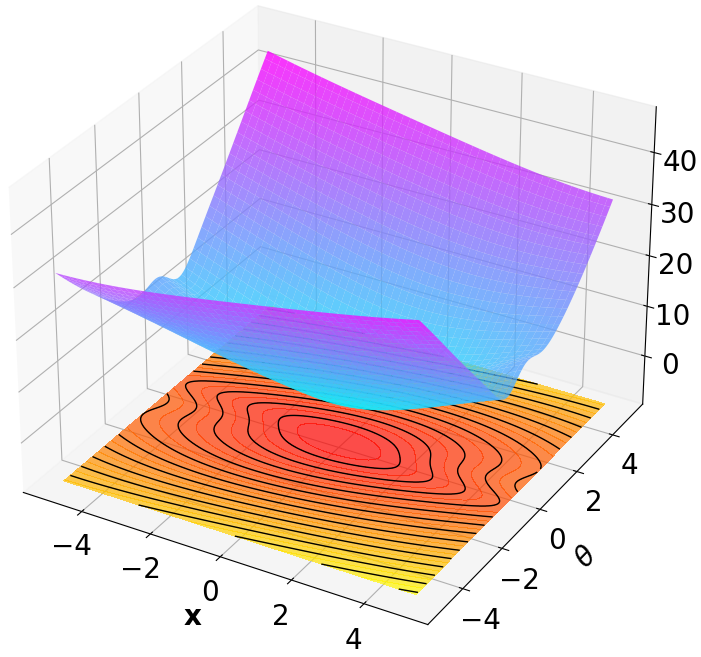}}
    \vspace{-5pt}
    \subfloat[]{\includegraphics[width=0.19\textwidth]{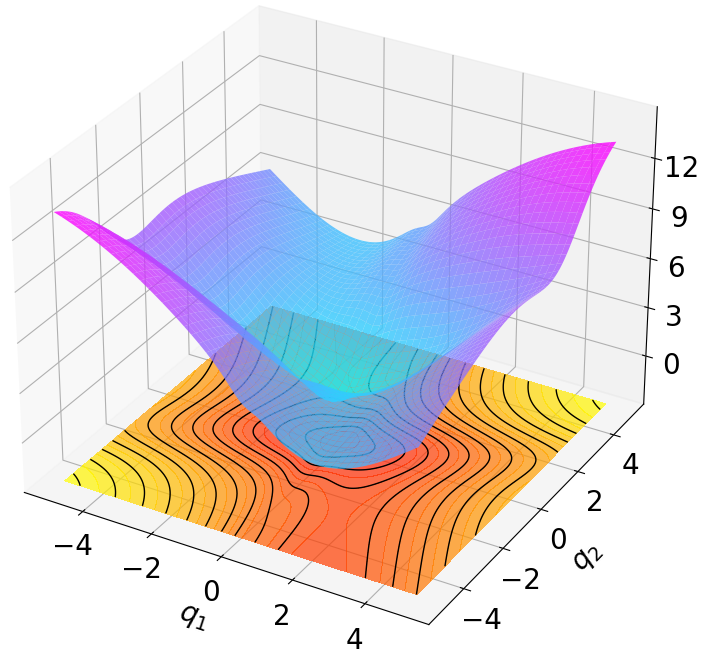}}\hfill
    \subfloat[]{\includegraphics[width=0.19\textwidth]{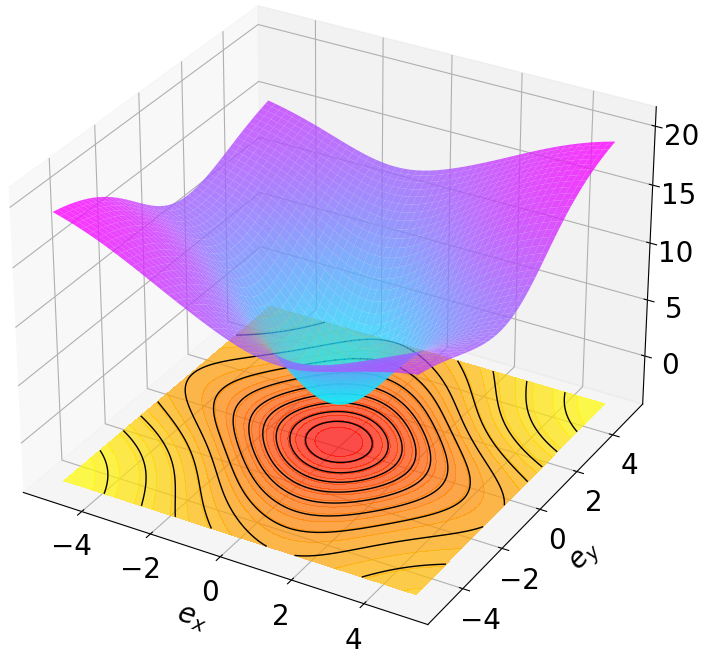}}\hfill
    \subfloat[]{\includegraphics[width=0.19\textwidth]{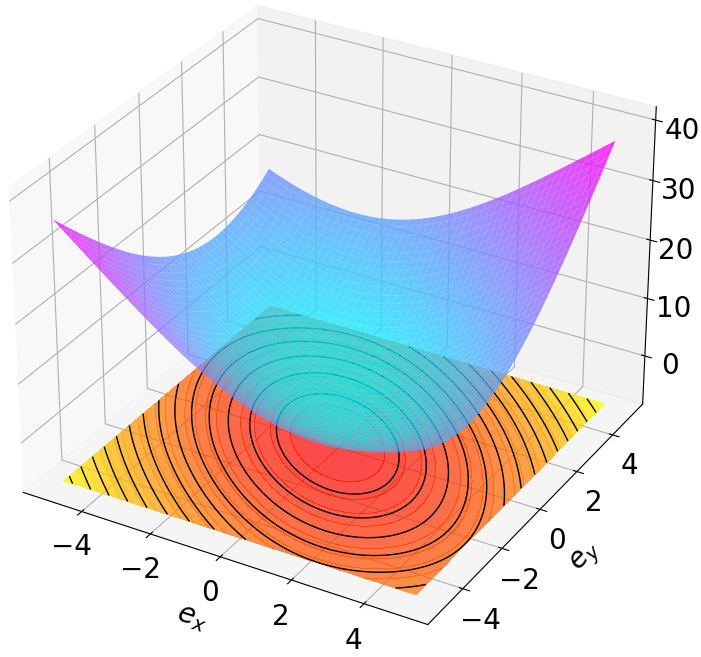}}\hfill
    \subfloat[]{\includegraphics[width=0.19\textwidth]{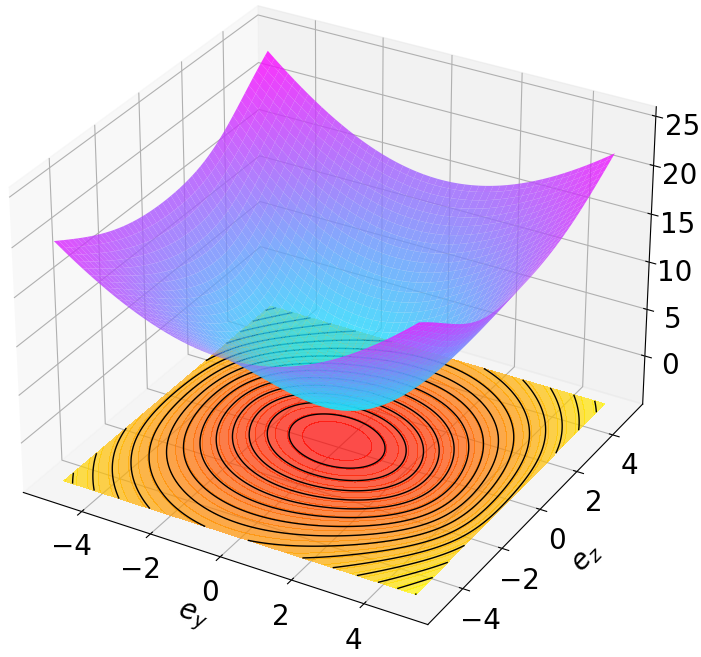}}\hfill
    \subfloat[]{\includegraphics[width=0.19\textwidth]{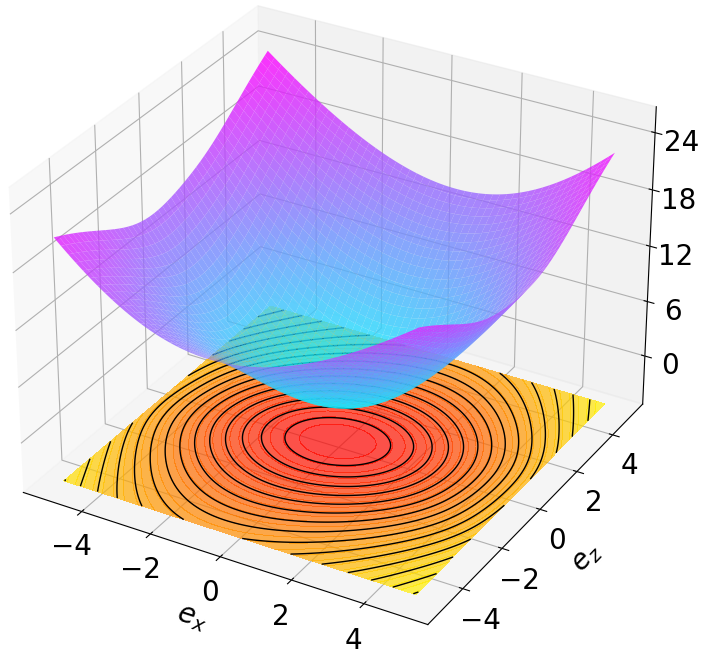}}
    \vspace{-5pt}
    \caption{Visualization of learned Lyapunov certificates and their corresponding contour plots across six benchmarks: (a) VanderPol; (b) Pendulum; (c)--(e) DuctedFan (projections in the $(x,y)$-plane with $\theta=0$, $(y,\theta)$-plane with $x=0$, and $(x,\theta)$-plane with $y=0$); (f) Two-link; (g) SingleCarTracking; (h)--(j) QuadrotorTracking (projections in the $(e_x,e_y)$-plane with $e_z=0$, $(e_y,e_z)$-plane with $e_x=0$, and $(e_x,e_z)$-plane with $e_y=0$).}
    \label{lyapunov_figs}
\end{figure*}

Finally, we visualize the Lyapunov networks $V_\psi$ for the optimal policies. The learned certificates and associated contour plots are shown in Fig. \ref{lyapunov_figs}. For high-dimensional systems, the plots illustrate the relationship between $V_\psi$ and two specific state components with all other components set to zero.

\begin{remark}\label{remark4}
    Since MSACL builds on the SAC framework, SAC serves as the primary baseline for our ablation study. As shown in Fig. \ref{training_reward_cost} and Table \ref{algorithm_eval_metrics}, the integration of the Lyapunov certificate enhances both training performance and stability metrics compared to the standard SAC. This validates that embedding Lyapunov-based constraints into MERL effectively yields neural policies with stability properties for model-free control systems.
\end{remark}

\subsection{Evaluation of Robustness and Generalization}
Neural policies with high-dimensional parameter spaces are prone to overfitting, often resulting in reduced robustness to noise and parametric variations, and limited generalization to unseen scenarios. This subsection evaluates the robustness and generalization of MSACL-trained policies, with the aim of demonstrating their real-world applicability under inherent uncertainties. 

For stabilizing tasks (VanderPol, Pendulum, DuctedFan, Two-link), we introduce process noise $\boldsymbol{\epsilon}_t \sim \mathcal{N}(\mathbf{0}, \sigma_{\text{env}} \cdot \mathbf{1}_d)$ to the system dynamics, where $\mathbf{1}_d$ denotes the all-ones vector of dimension $d$, yielding the perturbed form:
\begin{equation*}
    \mathbf{x}_{t+1} = f(\mathbf{x}_t, \mathbf{u}_t) + \boldsymbol{\epsilon}_t.
\end{equation*}
Additionally, we modify key physical parameters of the systems to assess whether the MSACL-trained policies can maintain stability under environmental variations.
For tracking tasks (SingleCarTracking and QuadrotorTracking), robustness is evaluated via dynamical noise, while generalization is assessed using reference trajectories $\mathbf{x}_t^{\text{ref}}$ that are significantly different from those in training. Table \ref{robust_para} details the specific parameter modifications and noise magnitudes.

\begin{table}[!ht]
    \renewcommand{\arraystretch}{1.0}
    \caption{System Parameters and Noise Magnitudes}
    \label{robust_para}
    \centering
    \begin{tabular}{ccc}
        \toprule
        Benchmarks           & System Parameters & Noise Magnitude ($\sigma_{\mathrm{env}}$) \\
        \midrule
        VanderPol            & $\mu = 0.5, 1.5$         & $0.5, 0.8$ \\
        Pendulum             & $L = 0.5, 1.0$           & $0.5, 1.0$   \\
        DuctedFan            & $m = 5, 12$              & $0.1, 0.3$   \\
        Two-link             & $l_1 = 0.75, l_2 = 1.5$  & $0.1, 0.2$   \\
        SingleCarTracking    & --                       & $0.1, 0.3$   \\
        QuadrotorTracking    & --                       & $0.01, 0.02$ \\
        \bottomrule
    \end{tabular}
\end{table}

\begin{figure*}[!htb]
    \centering
    \subfloat[\label{robust-1}]{\includegraphics[width=0.23\textwidth]{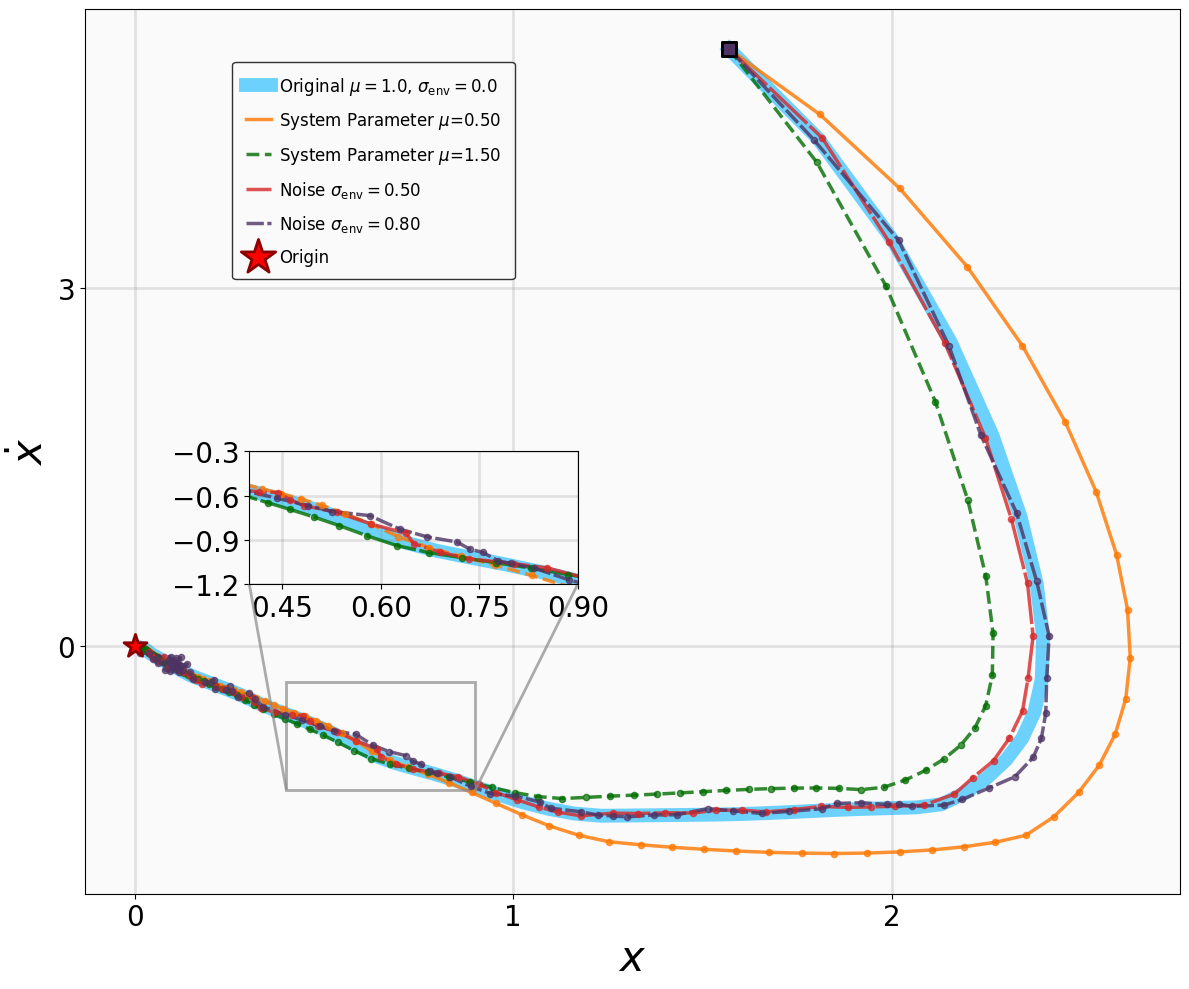}}%
    \hspace{0.01\textwidth}
    \subfloat[\label{robust-2}]{\includegraphics[width=0.23\textwidth]{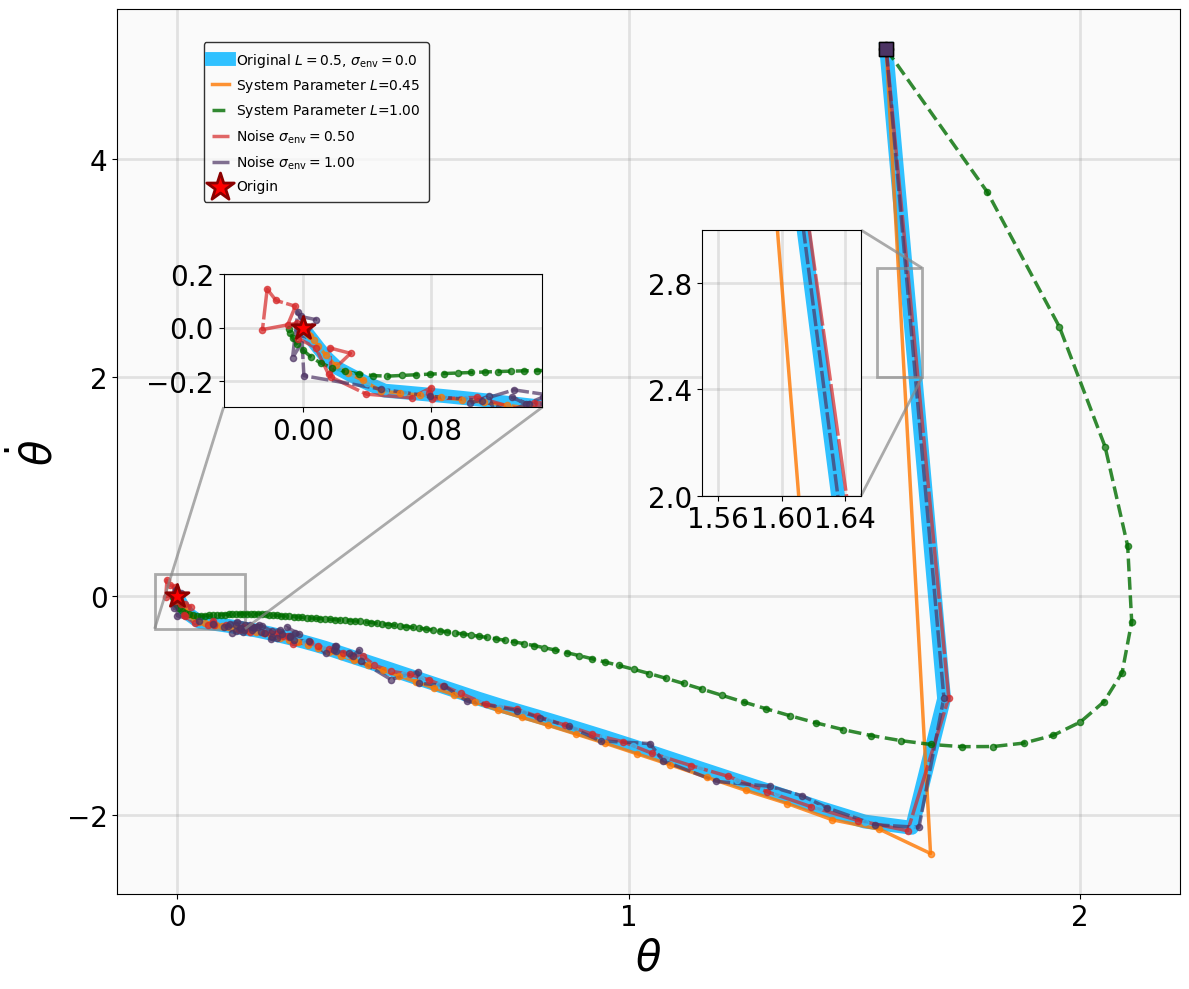}}%
    \hspace{0.01\textwidth}
    \subfloat[\label{robust-3}]{\includegraphics[width=0.23\textwidth]{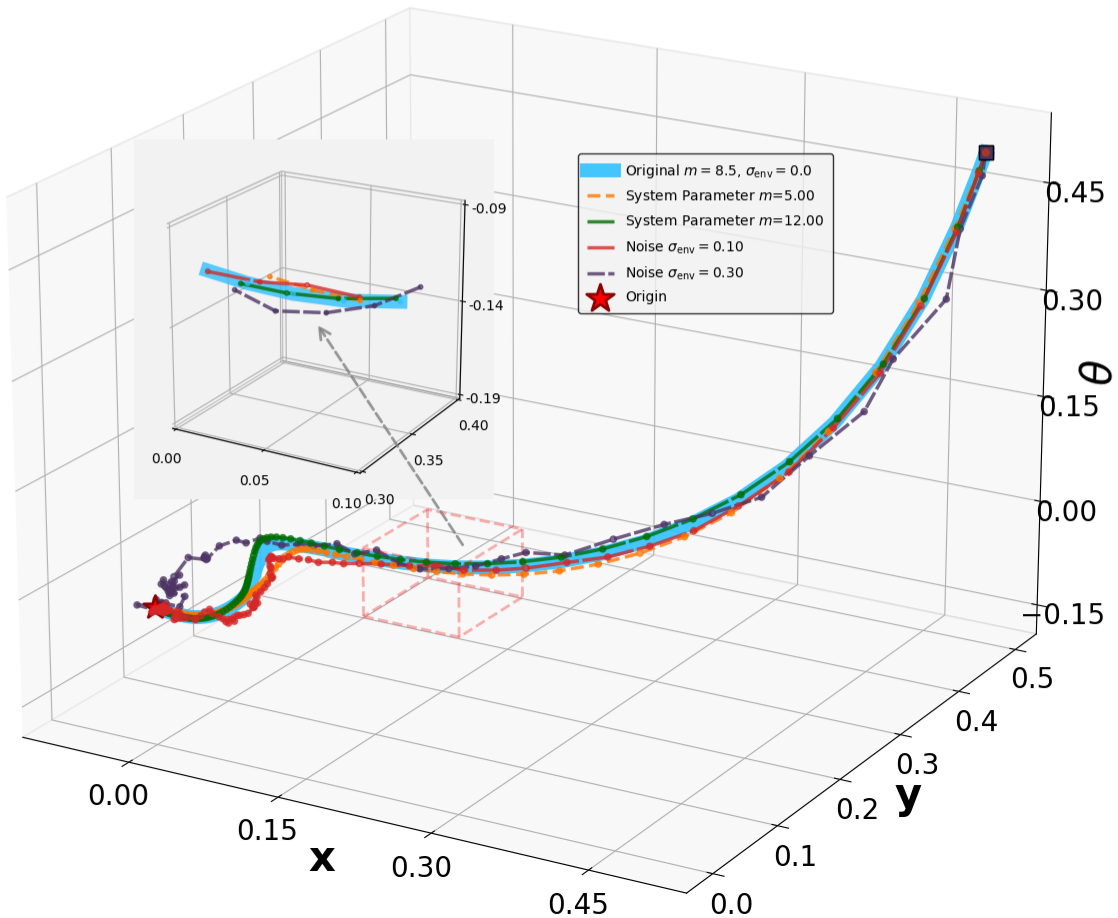}}%
    \hspace{0.01\textwidth}
    \subfloat[\label{robust-4}]{\includegraphics[width=0.23\textwidth]{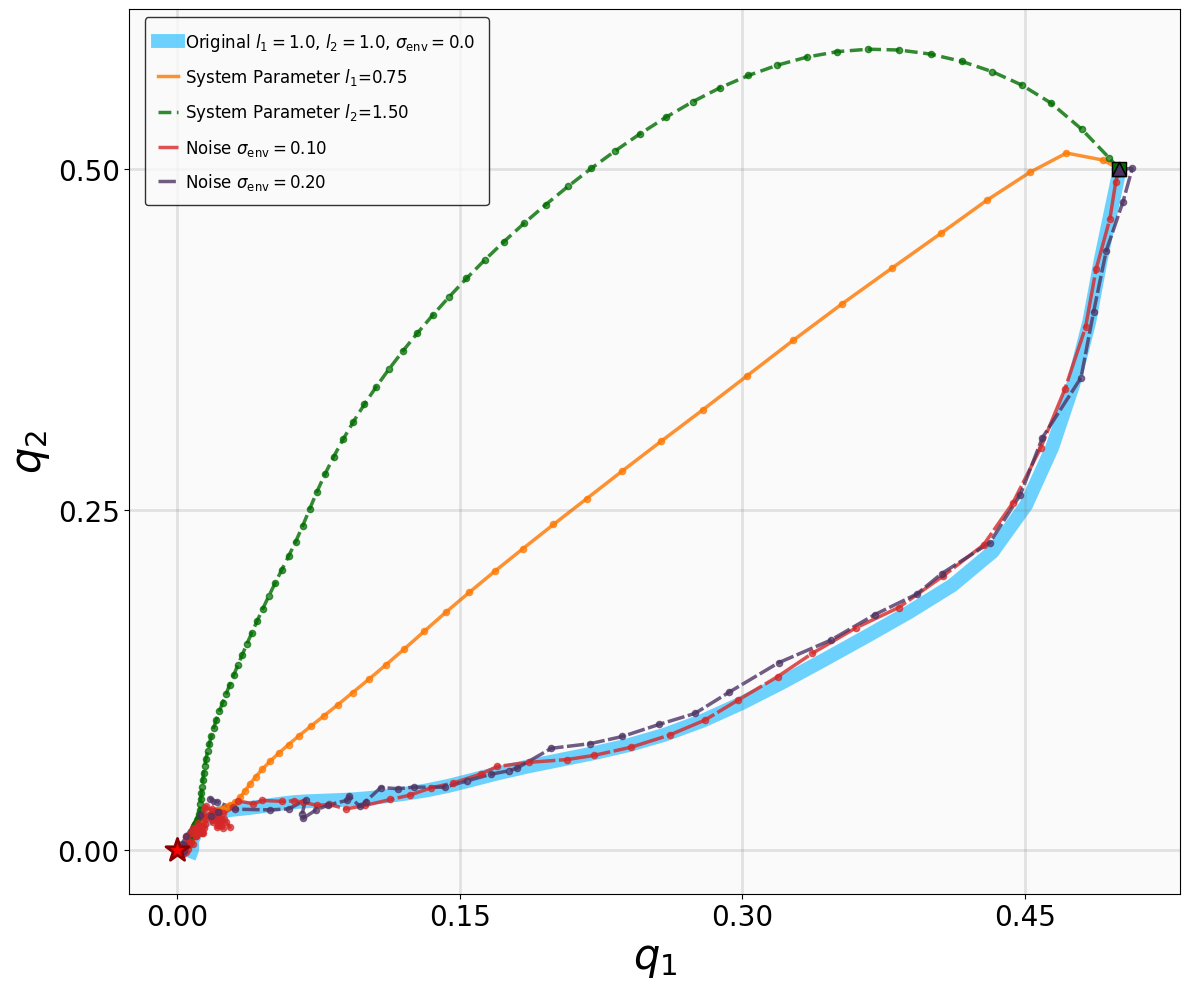}}%
    \vspace{-6pt}
    \caption{System trajectories under the MSACL-trained policies in stabilizing tasks subjected to environmental parametric uncertainties and process noise: (a) VanderPol with state variables $\mathbf{x} = [x, \dot{x}]^\top$; (b) Pendulum with angular position $\theta$ and angular velocity $\dot{\theta}$; (c) DuctedFan with coordinates $x, y$ and orientation $\theta$; (d) Two-link with joint angles $q_1$ and $q_2$. All trajectories exhibit robust convergence to the neighborhood of the equilibrium $\mathbf{x}_g = \mathbf{0}$.}
    \label{robust1-4}
\end{figure*}

\begin{figure*}[!ht]
    \centering
    \subfloat[\label{robust-STCar-1}]{\includegraphics[width=0.20\textwidth]{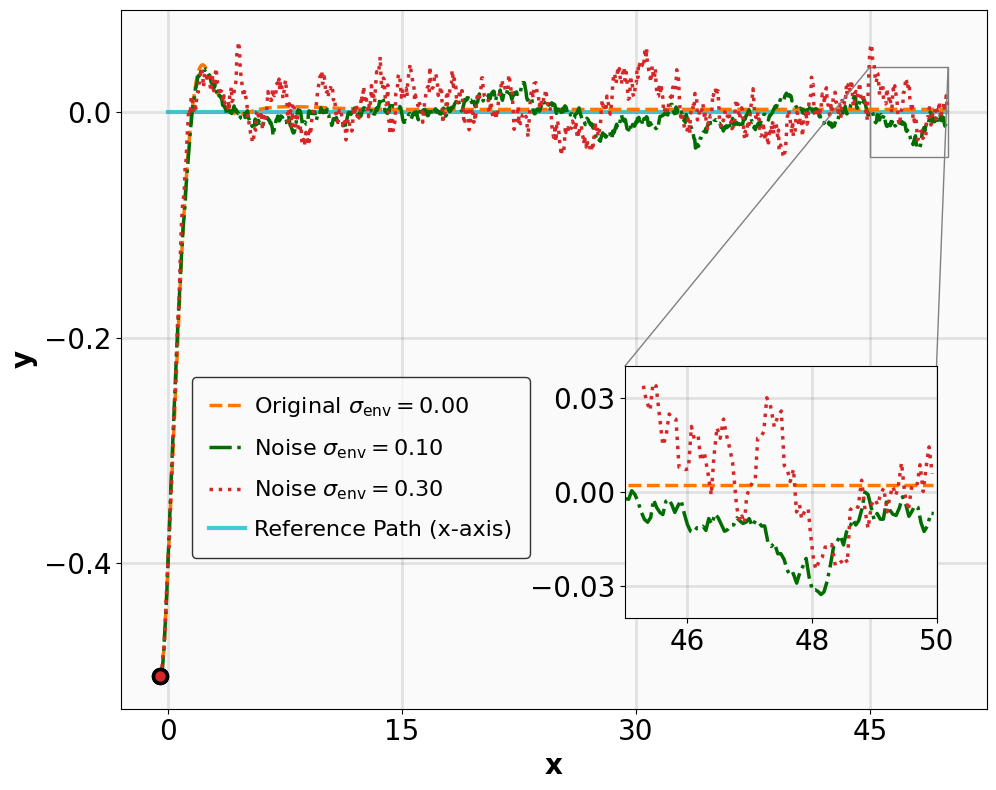}}\hfill
    \subfloat[\label{robust-STCar-2}]{\includegraphics[width=0.20\textwidth]{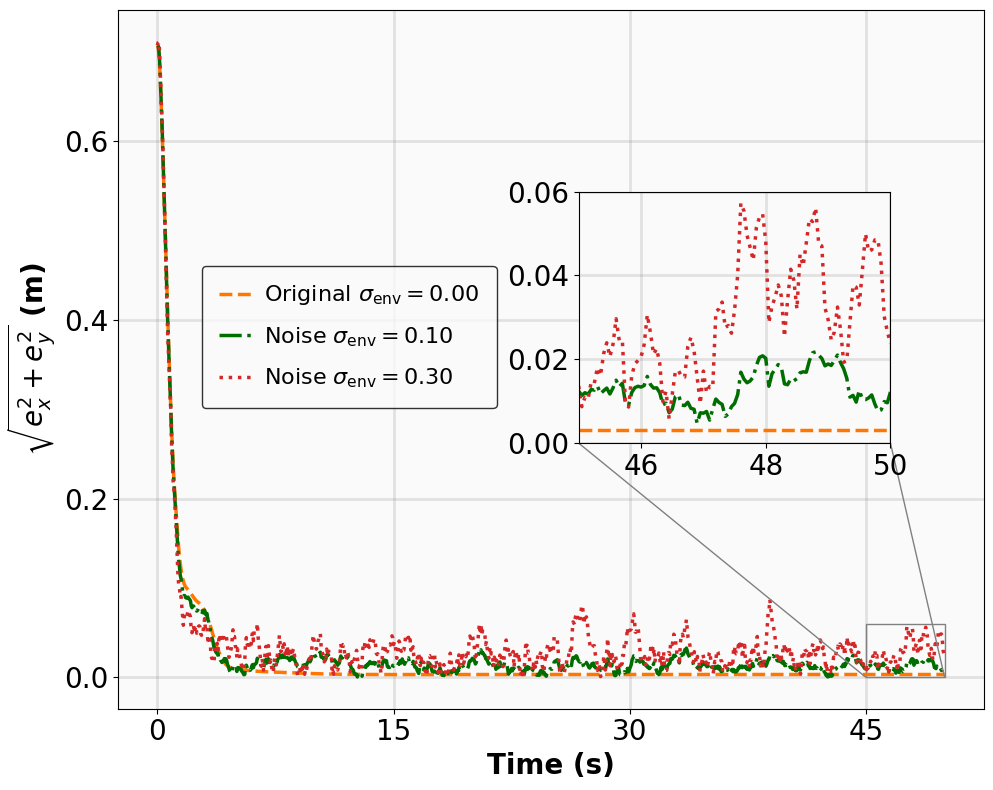}}\hfill
    \subfloat[\label{general-STCar-1}]{\includegraphics[width=0.175\textwidth]{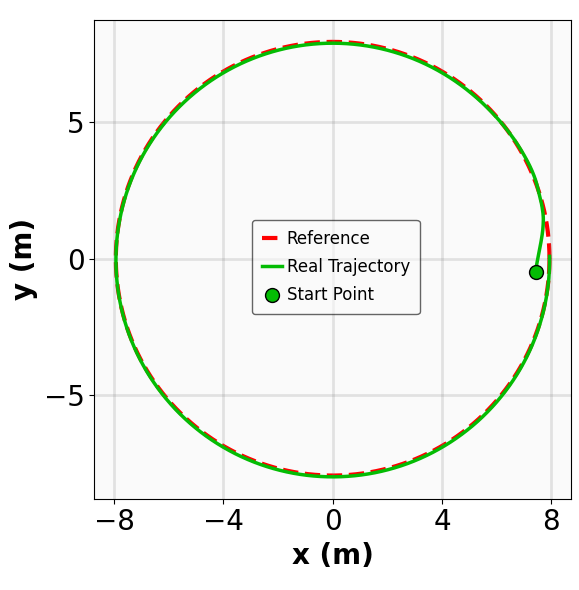}}\hfill
    \subfloat[\label{general-STCar-2}]{\includegraphics[width=0.17\textwidth]{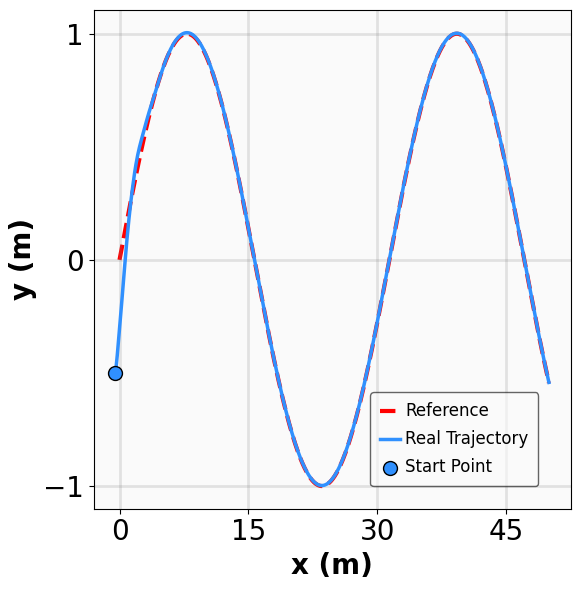}}\hfill
    \subfloat[\label{general-STCar-3}]{\includegraphics[width=0.18\textwidth]{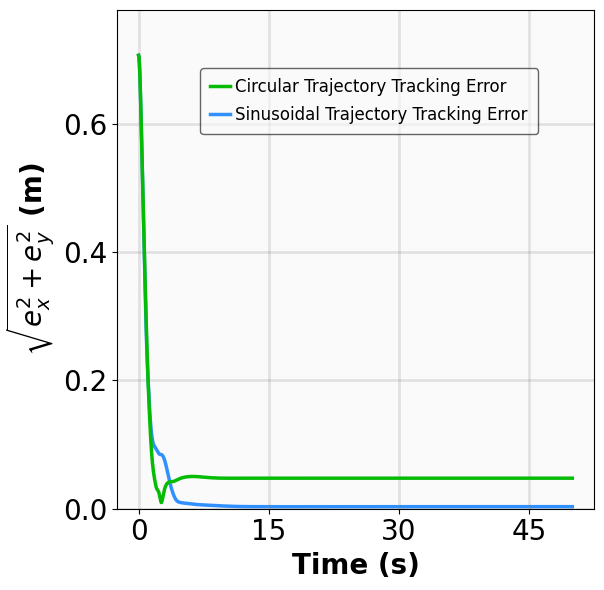}}\hfill
    \vspace{-6pt}
    \caption{Robustness (a)--(b) and generalization (c)--(e) evaluation for the SingleCarTracking task: (a) Car trajectories versus the straight-line reference $\mathbf{x}_t^{\text{ref}} = \{[t, 0] \in \mathbb{R}^2 \mid t\in [0, 50] \}$ under varying noise levels; (b) Evolution of the Euclidean position error $e_{\mathrm{xy}} = \sqrt{e_x^2 + e_y^2}$ corresponding to (a); (c) Tracking performance for a circular reference trajectory $\{[8\cos(0.125t),8\sin(0.125t)]^\top \in \mathbb{R}^2 \mid t\in [0, 50]\}$ using the policy trained on straight lines; (d) Tracking performance for a sinusoidal reference trajectory $\{[t,\sin(0.2t)]^\top \in \mathbb{R}^2 \mid t\in [0, 50]\}$; (e) Evolution of the position error $e_{\mathrm{xy}}$ for the circular and sinusoidal cases shown in (c) and (d). The results demonstrate robust convergence and effective generalization to unseen path geometries.} 
    \label{STCar1-5} 
\end{figure*}

\begin{figure*}[!ht]
    \centering
    \subfloat[\label{robust-Quad-1}]{\includegraphics[width=0.19\textwidth]{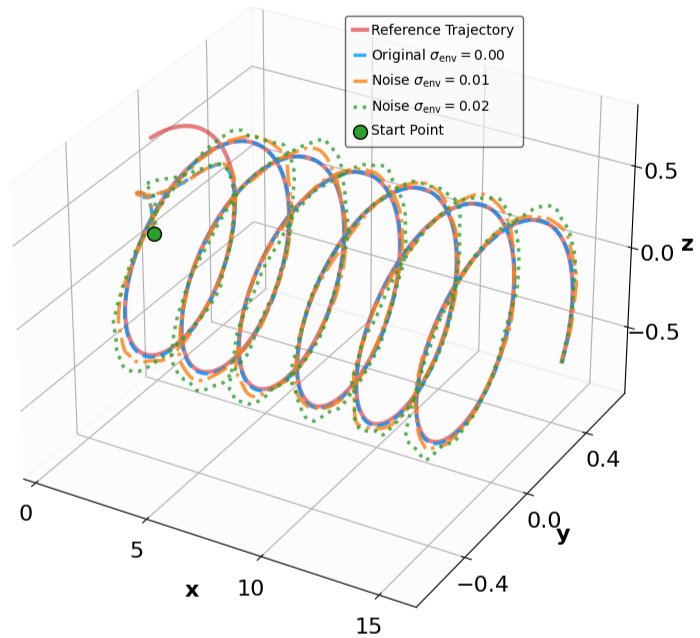}}\hfill
    \subfloat[\label{robust-Quad-2}]{\includegraphics[width=0.20\textwidth]{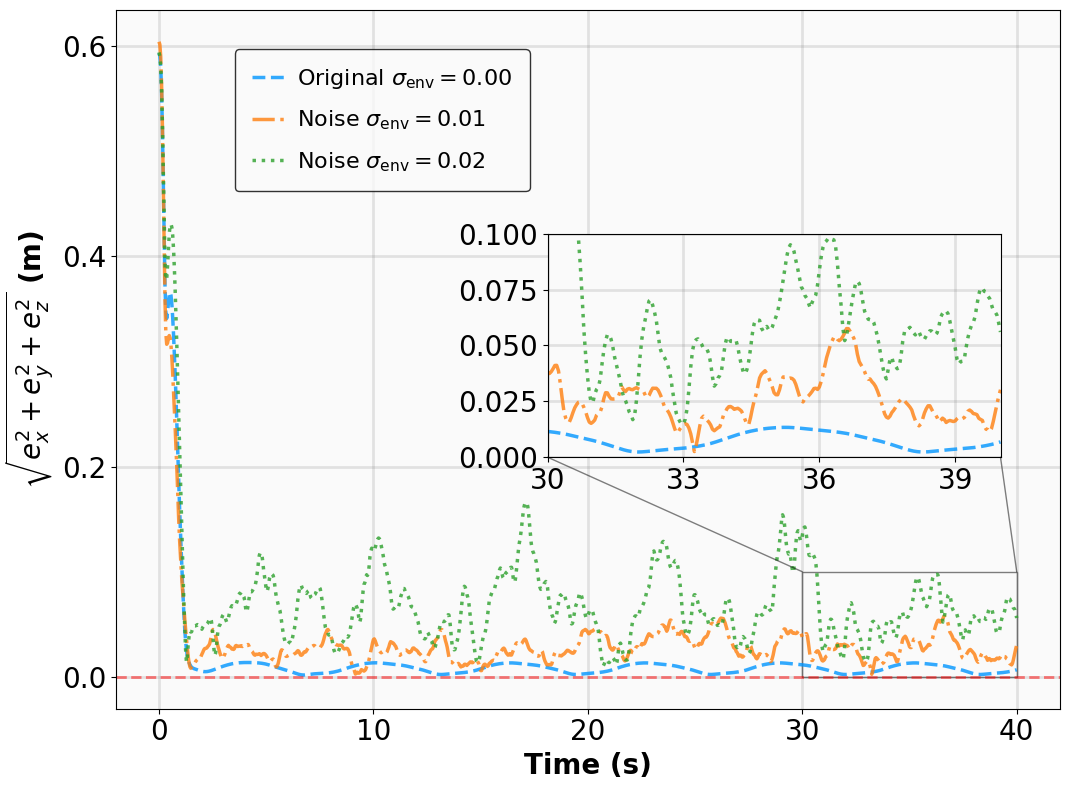}}\hfill
    \subfloat[\label{general-Quad-1}]{\includegraphics[width=0.19\textwidth]{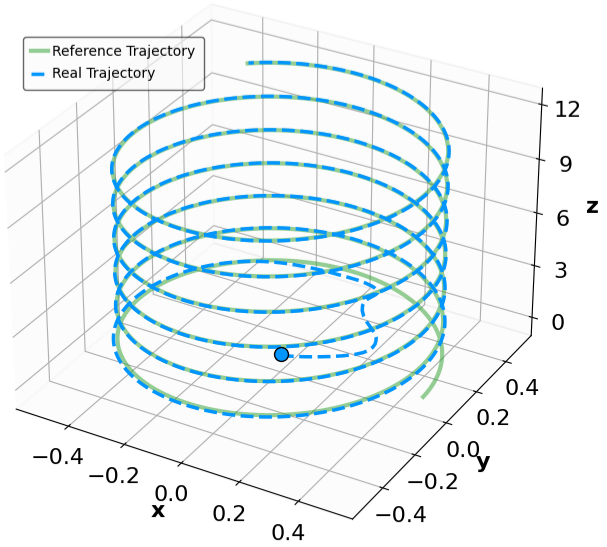}}\hfill
    \subfloat[\label{general-Quad-2}]{\includegraphics[width=0.19\textwidth]{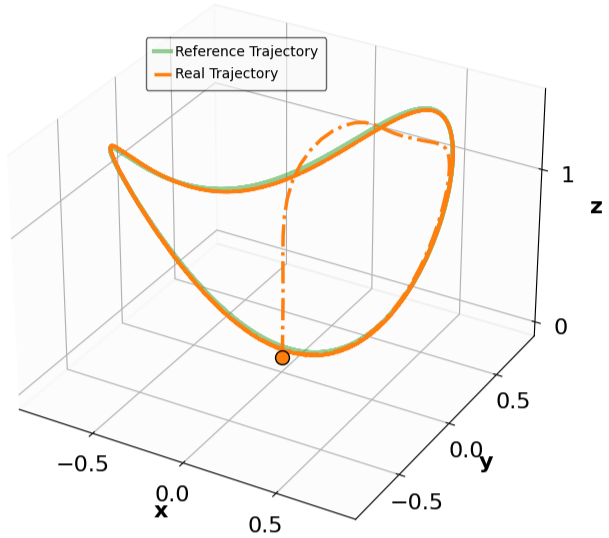}}\hfill
    \subfloat[\label{general-Quad-3}]{\includegraphics[width=0.20\textwidth]{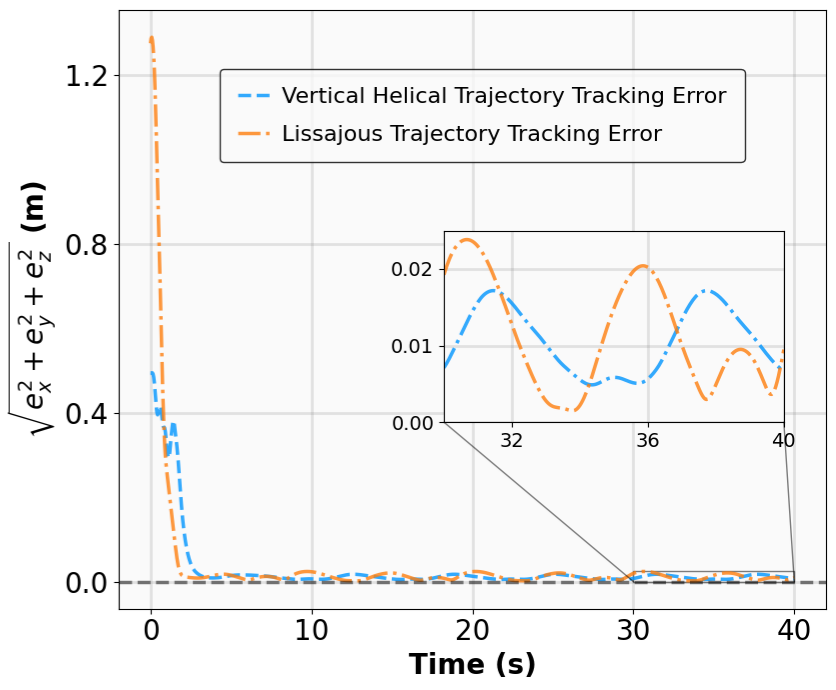}}\hfill
    \vspace{-6pt}
    \caption{Robustness (a)--(b) and generalization (c)--(e) evaluation for the QuadrotorTracking task: (a) Quadrotor trajectories versus the horizontal helical reference $\mathbf{x}_t^{\text{ref}} = \{[0.4t, 0.4\sin t, 0.6\cos t]^\top \mid t \in [0,40] \}$ under varying noise levels; (b) Evolution of the position error $e_{\mathrm{xyz}} = \sqrt{e_x^2 + e_y^2 + e_z^2}$ corresponding to (a); (c) Tracking performance for a vertical helical reference $\{[0.5\cos t, 0.5\sin t, 0.3t]^\top \in \mathbb{R}^3 \mid t \in [0,40] \}$ using the policy trained on the horizontal helix; (d) Tracking performance for a Lissajous reference trajectory $\{[0.8\sin(0.4t), 0.8\cos(0.4t), 0.4\sin(1.2t) + 1.0]^\top \in \mathbb{R}^3 \mid t \in [0,40] \}$; (e) Evolution of the position error $e_{\mathrm{xyz}}$ for the cases in (c) and (d). The results highlight the policy's robust tracking capability under environmental noise and its effective generalization to geometrically distinct reference trajectories.} 
    \label{Quad1-5} 
\end{figure*}

Robustness results for stabilizing tasks in Fig. \ref{robust1-4} show that all trajectories converge to the equilibrium $\mathbf{x}_g = \mathbf{0}$ despite parametric perturbations and process noise. 
Tracking task results (Fig. \ref{STCar1-5}, \ref{Quad1-5}) demonstrate that MSACL-trained policies effectively mitigate process noise within reasonable ranges. For SingleCarTracking, though trained on straight-line references, the policy generalizes well to circular and sinusoidal trajectories (Fig. \ref{general-STCar-1}, \ref{general-STCar-2}), with position error $e_{\mathrm{xy}} = \sqrt{e_x^2 + e_y^2}$ remaining below 0.1 in steady state (Fig. \ref{general-STCar-3}). 
For QuadrotorTracking, though trained on horizontal helical trajectories, the policy successfully tracks unseen vertical helix and Lissajous curves (Fig. \ref{general-Quad-1}, \ref{general-Quad-2}), with tracking error $e_{\mathrm{xyz}} = \sqrt{e_x^2 + e_y^2 + e_z^2}$ stabilizing around 0.02 (Fig. \ref{general-Quad-3}), demonstrating superior generalization capability.

These results demonstrate that MSACL-trained policies exhibit robustness to model uncertainties and process noise, while generalizing effectively to unseen reference signals. However, performance degrades under extreme disturbances or significant deviations from the training distribution. In such cases, MSACL can be redeployed to retrain policies for stable, high-precision control in specific operational scenarios.

\subsection{Sensitivity Analysis of the Multi-Step Horizon}
This subsection examines the impact of multi-step horizon $n$ on MSACL performance across all benchmarks. Following the protocol in Section \ref{training_results}, we evaluate policies trained with varying $n$ using performance metrics (AMCR, AMCC) and stability metrics (RR, ARS, AHS). For each benchmark, we conduct 100 experiments from fixed random initial states to compute these metrics. As summarized in Table \ref{n_step_results}, performance generally improves with larger $n$, as a longer horizon enables the policy to better identify the stability direction, i.e., the descent of $V_\psi$. This reduces training bias and reinforces stability properties.

Empirically, $n=20$ delivers satisfactory performance across all benchmarks, serving as a versatile default. However, excessively large $n$ may degrade performance due to increased cumulative variance in training data, potentially distorting the optimization landscape as analyzed in Section \ref{sec:lyapunov_loss}. Therefore, selecting an appropriate $n$ is essential to balance the bias-variance trade-off. For the control tasks herein, $n=20$ provides consistently robust results; for better performance, fine-tuning around this value is recommended.

\begin{table*}[!ht]
    \centering
    \begin{threeparttable}
    \caption{Experimental Results of MSACL Algorithm with different multi-step horizon Parameters $n$}
    \label{n_step_results}
    \begin{tabular}{cc cc *{4}{wc{1.7cm}}}
        \toprule
        \multirow{2}{*}{\textbf{Benchmark}} & \multirow{2}{*}{\textbf{$n$ Values}} & \multirow{2}{*}{\textbf{AMCR}} & \multirow{2}{*}{\textbf{AMCC}} & \multicolumn{4}{c}{\textbf{RR\,/\,ARS\,/\,AHS at different radii}} \\
        \cmidrule(lr){5-8}
        & & & & 0.2 & 0.1 & 0.05 & 0.01 \\
        \midrule

        \multirow{5}{*}{VanderPol} 
        & $n=1$  & 30.2 $\pm$ 64.5 & 31.1 $\pm$ 29.5 & 1.0\,/\,56.3\,/\,944.6 & 1.0\,/\,61.9\,/\,939.0 & 1.0\,/\,68.9\,/\,932.0 & 1.0\,/\,88.2\,/\,912.7 \\
        & $n=5$  & 31.5 $\pm$ 64.5 & \textbf{30.9 $\pm$ 29.4} & 1.0\,/\,58.4\,/\,942.4 & 1.0\,/\,63.4\,/\,937.5 & 1.0\,/\,67.6\,/\,933.3 & 1.0\,/\,78.7\,/\,922.2 \\
        & $n=10$ & 32.4 $\pm$ 64.6 & \textbf{30.9 $\pm$ 29.4} & 1.0\,/\,57.9\,/\,942.9 & 1.0\,/\,61.4\,/\,939.5 & 1.0\,/\,64.1\,/\,936.8 & 1.0\,/\,68.9\,/\,932.0 \\
        & $n=15$ & 33.1 $\pm$ 64.4 & 31.2 $\pm$ 29.5 & \textbf{1.0\,/\,54.1\,/\,946.7} & \textbf{1.0\,/\,55.1\,/\,945.8} & \textbf{1.0\,/\,55.7\,/\,945.3} & \textbf{1.0\,/\,57.0\,/\,943.9} \\
        & $n=20$ & \textbf{33.2 $\pm$ 64.5} & 31.0 $\pm$ 29.5 & 1.0\,/\,57.8\,/\,943.1 & 1.0\,/\,58.7\,/\,942.2 & 1.0\,/\,59.2\,/\,941.7 & 1.0\,/\,59.5\,/\,941.4 \\
        \cmidrule(lr){1-8}
        
        \multirow{5}{*}{Pendulum} 
        & $n=1$  & -578.5 $\pm$ 1232.9 & 444.2 $\pm$ 815.1 & 0.77\,/\,37.1\,/\,963.5 & 0.77\,/\,42.4\,/\,958.6 & 0.77\,/\,44.4\,/\,956.5 & 0.77\,/\,48.8\,/\,952.1 \\
        & $n=5$  & 34.7 $\pm$ 360.0 & 30.7 $\pm$ 171.8 & 0.98\,/\,36.8\,/\,964.0 & 0.98\,/\,44.7\,/\,956.2 & 0.98\,/\,46.0\,/\,954.9 & 0.98\,/\,49.4\,/\,951.5 \\
        & $n=10$ & 35.7 $\pm$ 360.1 & 31.0 $\pm$ 171.7 & \textbf{0.98\,/\,34.9\,/\,966.0} & \textbf{0.98\,/\,35.6\,/\,965.4} & \textbf{0.98\,/\,36.1\,/\,964.9} & \textbf{0.98\,/\,38.5\,/\,962.3} \\
        & $n=15$ & 35.2 $\pm$ 360.0 & 31.0 $\pm$ 171.7 & 0.98\,/\,35.2\,/\,965.6 & 0.98\,/\,36.9\,/\,964.0 & 0.98\,/\,38.4\,/\,962.6 & 0.98\,/\,44.0\,/\,956.9 \\
        & $n=20$ & \textbf{35.8 $\pm$ 355.7} & \textbf{30.3 $\pm$ 167.9} & 0.98\,/\,38.0\,/\,962.8 & 0.98\,/\,39.3\,/\,961.6 & 0.98\,/\,40.7\,/\,960.2 & 0.98\,/\,44.1\,/\,956.8 \\
        \cmidrule(lr){1-8}
        
        \multirow{5}{*}{Ducted Fan} 
        & $n=1$  & -116.3 $\pm$ 584.8 & 163.2 $\pm$ 463.3 & 0.89\,/\,32.3\,/\,967.8 & 0.89\,/\,49.8\,/\,950.6 & 0.89\,/\,57.0\,/\,943.9 & 0.89\,/\,83.5\,/\,917.3 \\
        & $n=5$  & 87.4 $\pm$ 3.7 & \textbf{2.0 $\pm$ 1.3} & 1.0\,/\,35.5\,/\,965.2 & \textbf{1.0\,/\,49.8\,/\,950.9} & \textbf{1.0\,/\,57.9\,/\,942.9} & 1.0\,/\,78.4\,/\,922.1 \\
        & $n=10$ & \textbf{88.6 $\pm$ 3.5} & 2.1 $\pm$ 1.3 & \textbf{1.0\,/\,35.3\,/\,965.3} & 1.0\,/\,52.7\,/\,948.0 & 1.0\,/\,59.6\,/\,941.3 & \textbf{1.0\,/\,68.9\,/\,932.0} \\
        & $n=15$ & 88.1 $\pm$ 4.0 & \textbf{2.0 $\pm$ 1.3} & 1.0\,/\,36.1\,/\,964.6 & 1.0\,/\,52.0\,/\,949.0 & 1.0\,/\,61.4\,/\,939.5 & 1.0\,/\,74.3\,/\,926.6 \\
        & $n=20$ & 73.7 $\pm$ 37.7 & 4.4 $\pm$ 8.5 & 0.95\,/\,56.3\,/\,944.2 & 0.95\,/\,79.2\,/\,921.6 & 0.95\,/\,94.5\,/\,906.4 & 0.95\,/\,127.0\,/\,873.9 \\
        \cmidrule(lr){1-8}
        
        \multirow{5}{*}{Two-link} 
        & $n=1$  & -208.6 $\pm$ 372.5 & 216.0 $\pm$ 267.0 & 0.60\,/\,13.8\,/\,987.2 & 0.60\,/\,22.3\,/\,976.7 & 0.60\,/\,29.6\,/\,971.3 & 0.60\,/\,40.4\,/\,960.5 \\
        & $n=5$  & -142.4 $\pm$ 340.0 & 167.6 $\pm$ 241.6 & 0.67\,/\,13.7\,/\,987.0 & 0.67\,/\,20.3\,/\,980.4 & 0.67\,/\,24.7\,/\,976.1 & 0.67\,/\,29.9\,/\,971.0 \\
        & $n=10$ & -163.8 $\pm$ 346.6 & 183.1 $\pm$ 247.1 & 0.64\,/\,10.7\,/\,990.2 & 0.64\,/\,16.4\,/\,984.5 & 0.64\,/\,23.2\,/\,977.6 & 0.64\,/\,37.0\,/\,963.9 \\
        & $n=15$ & -131.0 $\pm$ 354.1 & 158.3 $\pm$ 251.4 & 0.71\,/\,12.1\,/\,988.7 & 0.71\,/\,18.5\,/\,982.3 & 0.71\,/\,24.8\,/\,976.1 & 0.71\,/\,40.6\,/\,960.3 \\
        & $n=20$ & \textbf{89.8 $\pm$ 7.5} & \textbf{0.4 $\pm$ 0.2} & \textbf{1.0\,/\,18.1\,/\,982.0} & \textbf{1.0\,/\,25.0\,/\,975.8} & \textbf{1.0\,/\,28.6\,/\,972.2} & \textbf{1.0\,/\,35.2\,/\,965.7} \\
        \cmidrule(lr){1-8}
        
        \multirow{5}{*}{SingleCarTracking} 
        & $n=1$  & -51.0 $\pm$ 194.9 & 46.4 $\pm$ 62.6 & 0.64\,/\,32.0\,/\,964.0 & 0.64\,/\,64.6\,/\,931.1 & 0.64\,/\,79.7\,/\,921.2 & 0.64\,/\,92.1\,/\,908.8 \\
        & $n=5$  & -40.2 $\pm$ 217.9 & 35.0 $\pm$ 58.8 & 0.74\,/\,31.6\,/\,968.7 & 0.74\,/\,56.1\,/\,943.8 & 0.74\,/\,78.9\,/\,920.9 & 0.74\,/\,113.0\,/\,887.8 \\
        & $n=10$ & \textbf{84.7 $\pm$ 32.7} & \textbf{2.9 $\pm$ 14.1} & \textbf{0.99\,/\,37.6\,/\,961.6} & \textbf{0.99\,/\,55.3\,/\,940.7} & \textbf{0.99\,/\,76.5\,/\,924.1} & 0.99\,/\,96.4\,/\,904.5 \\
        & $n=15$ & 67.5 $\pm$ 84.7 & 8.8 $\pm$ 32.5 & 0.95\,/\,35.2\,/\,964.8 & 0.95\,/\,57.8\,/\,938.7 & 0.95\,/\,82.6\,/\,917.0 & 0.95\,/\,109.1\,/\,891.7 \\
        & $n=20$ & 82.6 $\pm$ 55.1 & 3.0 $\pm$ 15.6 & 0.99\,/\,37.6\,/\,960.5 & 0.99\,/\,63.1\,/\,934.3 & 0.99\,/\,85.0\,/\,915.8 & \textbf{0.99\,/\,95.0\,/\,905.6} \\
        \cmidrule(lr){1-8}
        
        \multirow{5}{*}{QuadrotorTracking} 
        & $n=1$  & -12708.9 $\pm$ 645.4 & 12582.7 $\pm$ 645.2 & 0.0\,/\,--\,/\,-- & 0.0\,/\,--\,/\,-- & 0.0\,/\,--\,/\,-- & 0.0\,/\,--\,/\,-- \\
        & $n=5$  & -4708.7 $\pm$ 3158.6 & 4646.3 $\pm$ 3135.9 & 0.0\,/\,--\,/\,-- & 0.0\,/\,--\,/\,-- & 0.0\,/\,--\,/\,-- & 0.0\,/\,--\,/\,-- \\
        & $n=10$ & 534.3 $\pm$ 5.0 & 11.8 $\pm$ 0.9 & 1.0\,/\,54.4\,/\,946.5 & 1.0\,/\,66.2\,/\,934.7 & 1.0\,/\,116.5\,/\,650.9 & 0.0\,/\,--\,/\,-- \\
        & $n=15$ & \textbf{858.5 $\pm$ 1.0} & \textbf{10.8 $\pm$ 0.5} & \textbf{1.0\,/\,31.3\,/\,969.6} & 1.0\,/\,36.8\,/\,964.1 & 1.0\,/\,41.2\,/\,959.7 & \textbf{1.0}\,/\,115.4\,/\,\textbf{747.9} \\
        & $n=20$ & 838.5 $\pm$ 1.5 & 14.7 $\pm$ 2.0 & 1.0\,/\,33.1\,/\,967.8 & \textbf{1.0\,/\,34.1\,/\,966.8} & \textbf{1.0\,/\,37.0\,/\,963.9} & \textbf{1.0\,/\,49.8}\,/\,620.0 \\
        
        \bottomrule
    \end{tabular}
    \end{threeparttable}
\end{table*}
\vspace{-10pt}

\section{Discussion and Conclusion}\label{sec:conclusion}
\subsection{Discussion}
This subsection elucidates the functional roles of MSACL components and discusses the fundamental limitations of our approach. Functionally, MSACL augments the SAC framework with a stability-oriented certificate, departing from standard RL that relies on exhaustive exploration to discover stabilizing behaviors. Instead, MSACL leverages a learned Lyapunov certificate to provide guidance for policy updates. By categorizing trajectory segments through proposed sampling mechanism, the exponential stability constraints are coupled with the training samples, resulting in an empirically accurate Lyapunov certificate. Consequently, the stability advantage explicitly prioritizes actions that facilitate Lyapunov descent, effectively guiding the policy toward the region of stability.

Despite its favorable empirical performance, particularly its strong effectiviness and high efficiency, it is crucial to acknowledge that MSACL, fundamentally rooted in model-free RL, lacks formal theoretical guarantees for system stability. In contrast, traditional frameworks such as cMDPs \cite{altman1999constrained, achiam2017constrained} and Hamilton-Jacobi reachability analysis \cite{bansal2017hamilton, fisac2019bridging} offer rigorous mathematical formulations that provide strict theoretical guarantees for system behavior. Nevertheless, these formal methods often struggle with the curse of dimensionality and high computational costs in unknown environments. Bridging the gap between the empirical scalability of deep RL methods like MSACL and the rigorous theoretical guarantees of formal mathematical frameworks remains an essential challenge for future research.

\subsection{Conclusion}
We in this article presented MSACL, a multi-step Lyapunov-guided reinforcement learning algorithm that provides an efficient framework for stabilizing control tasks. By integrating a Lyapunov-oriented critic into an off-policy MERL architecture, MSACL enables rapid and robust stabilization for model-free systems while improving sample efficiency through data reuse and broader exploration. Extensive experiments across multiple benchmarks demonstrated the consistent empirical superiority of MSACL over established baselines in terms of performance, robustness, and generalization capabilities. 

Despite these promising results, the learned Lyapunov certificates in MSACL remain empirical approximations and do not yet provide strict formal guarantees comparable to those of classical analytical methods. Therefore, an important direction for future work is to integrate MSACL with formal verification tools. In particular, we plan to combine the proposed multi-step RL framework with structured analytical Lyapunov certificate templates, with the goal of enhancing both theoretical rigor and practical reliability. By bridging the scalability and exploration efficiency of MSACL with the mathematical rigor of formal methods, we hope to develop learning-based control strategies that are not only highly efficient, but also equipped with provable stability and safety guarantees.

\end{document}